\documentclass[sigconf]{acmart}
\usepackage[utf8]{inputenc}
\usepackage{mathtools}
\usepackage{hyperref}
\usepackage{booktabs}
\usepackage{algorithm}
\usepackage{algorithmic}
\usepackage{bm}
\usepackage{multirow}
\usepackage{enumitem}
\usepackage{amsmath,amsfonts,bm}

\def\vone{{\bm{1}}}

\def\sR{{\mathbb{R}}}

\def\gL{{\mathcal{L}}}
\def\gM{{\mathcal{M}}}
\def\gO{{\mathcal{O}}}
\def\gP{{\mathcal{P}}}
\def\gQ{{\mathcal{Q}}}
\def\gD{{\mathcal{D}}}
\def\gG{{\mathcal{G}}}
\def\gS{{\mathcal{S}}}

\def\rva{{\mathbf{a}}}
\def\rvb{{\mathbf{b}}}
\def\rvf{{\mathbf{f}}}
\def\rvg{{\mathbf{g}}}
\def\rvu{{\mathbf{u}}}
\def\rvv{{\mathbf{v}}}
\def\rvx{{\mathbf{x}}}
\def\rvy{{\mathbf{y}}}

\def\rmA{{\mathbf{A}}}
\def\rmC{{\mathbf{C}}}
\def\rmK{{\mathbf{K}}}
\def\rmN{{\mathbf{N}}}

\def\rmP{{\mathbf{P}}}
\def\rmT{{\mathbf{T}}}
\def\rmX{{\mathbf{X}}}

\DeclareMathOperator*{\argmin}{arg\,min}
\DeclareMathOperator*{\argmax}{arg\,max}

\DeclareMathOperator{\Exposure}{Exposure}

\newcommand{\transpose}{^\mathsf{T}}
\newcommand{\Gumbel}{\text{Gumbel}}

\newcommand{\answerTODO}[1][]{\textcolor{red}{\bf [TODO]}}
\newcommand\rqone{\textbf{RQ1}}
\newcommand\rqtwo{\textbf{RQ2}}
\newcommand\rqthree{\textbf{RQ3}}

\newcommand{\MOT}{\textsf{MOT}}
\newcommand{\CoMOT}{\textsf{CoMOT}}
\newcommand{\fair}{\text{fair}}

\DeclareMathOperator{\grad}{grad}
\DeclareMathOperator{\AutoDiff}{AutoDiff}
\DeclareMathOperator{\AdamOperator}{Adam}

\AtBeginDocument{%
  \providecommand\BibTeX{{%
    \normalfont B\kern-0.5em{\scshape i\kern-0.25em b}\kern-0.8em\TeX}}}

\setcopyright{acmcopyright}
\copyrightyear{2023}
\acmYear{2023}
\setcopyright{acmlicensed}\acmConference[SIGIR '23]{Proceedings of the 46th International ACM SIGIR Conference on Research and Development in Information Retrieval}{July 23--27, 2023}{Taipei, Taiwan}
\acmBooktitle{Proceedings of the 46th International ACM SIGIR Conference on Research and Development in Information Retrieval (SIGIR '23), July 23--27, 2023, Taipei, Taiwan}
\acmPrice{15.00}
\acmDOI{10.1145/3539618.3591714}
\acmISBN{978-1-4503-9408-6/23/07}

\begin{document}

%%% check references
%\nocite{*}

\title{Learning to Re-rank with Constrained Meta-Optimal Transport}

\author{Andrés Hoyos-Idrobo}
\email{andres.hoyosidrobo@rakuten.com}
\orcid{0000-0003-1729-1927}
\affiliation{%
  \institution{Rakuten Institute of Technology, Rakuten Group, Inc.}
  %\streetaddress{92 Rue Reaumur}
  %\city{Paris}
  %\state{{\^}Ile de France}
  \country{France}
  %\postcode{75002}
}



\begin{abstract}
Many re-ranking strategies in search systems rely on stochastic ranking policies, encoded as Doubly-Stochastic (DS) matrices, that satisfy desired ranking constraints in expectation, e.g., Fairness of Exposure (FOE). These strategies are generally two-stage pipelines: \emph{i)} an offline re-ranking policy construction step and \emph{ii)} an online sampling of rankings step. 
Building a re-ranking policy requires repeatedly solving a constrained optimization problem, one for each issued query. Thus, it is necessary to recompute the optimization procedure for any new/unseen query. 
Regarding sampling, the Birkhoff-von-Neumann decomposition (BvND) is the favored approach to draw rankings from any DS-based policy. Nonetheless, the BvND is too costly to compute online. Hence, the BvND as a sampling solution is memory-consuming as it can grow as $\gO(N\, n^2)$ for $N$ queries and $n$ documents.  

This paper proposes a novel, fast, lightweight way to predict fair stochastic re-ranking policies: Constrained Meta-Optimal Transport (CoMOT). This method fits a neural network shared across queries like a learning-to-rank system. We also introduce Gumbel-Matching Sampling (GumMS), an online sampling approach from DS-based policies. Our proposed pipeline, CoMOT + GumMS, only needs to store the parameters of a single model, and it can generalize to unseen queries. We empirically evaluated our pipeline on the TREC 2019 and 2020 datasets under FOE constraints. Our experiments show that CoMOT rapidly predicts fair re-ranking policies on held-out data, with a speed-up proportional to the average number of documents per query. It also displays fairness and ranking performance similar to the original optimization-based policy. Furthermore, we empirically validate the effectiveness of GumMS to approximate DS-based policies in expectation. Together, our methods are an important step in learning-to-predict solutions to optimization problems in information retrieval.

\end{abstract}

\begin{CCSXML}
<ccs2012>
	<concept>
    	<concept_id>10002951.10003317.10003338.10003343</concept_id>
    	<concept_desc>Information systems~Learning to rank</concept_desc>
    	<concept_significance>500</concept_significance>
	</concept>
</ccs2012>
\end{CCSXML}
	
\ccsdesc[500]{Information systems~Learning to rank}

\keywords{fairness in rankings, Fairness of exposure, optimal transport}


\maketitle

\section{Introduction}
Ranking systems are ubiquitous and prevalent in our daily lives. These systems are the core of many online marketplaces, job search engines, and media streaming platforms. Therefore, they also directly or indirectly impact our lives. For example, ranking systems influence item discovery, consumption, and purchase. Generally, these systems aim to maximize short-term engagement, item purchase, or other indicators depending on business needs, i.e., utility. However, there is a growing body of evidence ~\cite{kay2015unequal, diaz2020evaluating} that information retrieval (IR) methods that focus only on maximizing the ranking utility may disparately expose items of similar relevance from the collection~\cite{zehlike2017fa,asudeh2019designing,celis2017ranking,biega2018equity}. Such disparities in exposure-outcome raise concerns of algorithmic fairness and bias of moral import~\cite{friedman1996bias}, as they may contribute to representational and allocative harms~\cite{crawford2017trouble}. For instance, some creators may have an advantage in the system based on their prior popularity~\cite{abdollahpouri2019popularity}. Similarly, the ranking system may reproduce historical and ongoing social discrimination by disadvantaging creators of a particular gender or race.

Re-ranking policies are post hoc approaches to modify the behavior of ranking systems while keeping the ranker model fixed. These policies aim to optimize simultaneously multiple objectives, e.g., fairness and utility. Regarding fairness, the IR research community concentrates much of its effort on imposing socioeconomic constraints, e.g., social welfare~\cite{su2022optimizing},  Generalized Gini welfare Functions~\cite{do2022optimizing}, or allocating exposure~\cite{singh2021fairness, singh2018fairness}. However, these constraints balance fairness and utility in expectation. Thus, a stochastic ranking policy is the most natural way to enforce them. 

\begin{figure}[!ht]
	\small
	\centering
	{\includegraphics[width=1\linewidth, trim={0 0mm 0 0mm}, clip]{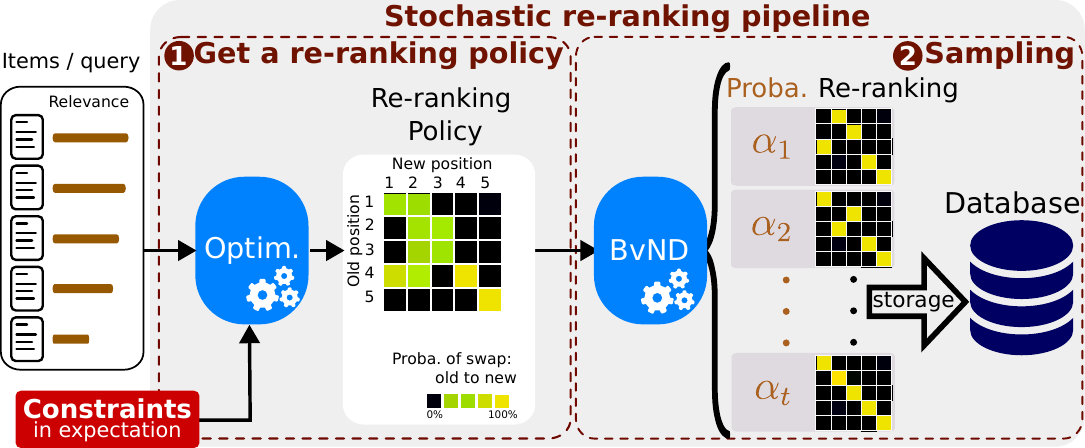}}
\caption{Illustration of the constrained stochastic re-ranking problem.
We obtain a fair probabilistic re-ranking policy for each query by solving a constrained optimization problem. Then, we decompose each policy to get many rankings with their corresponding occurrence probabilities and store the tuple (query, probability, ranking). During deployment, we retrieve a ranking using its associated probability.
}
	\label{fig:problem}
	\Description{
    Current fair re-ranking methods do not leverage the ranking data in the optimization stage. Also, these methods are currently memory inefficient.
    }
\end{figure}

So far, the literature on stochastic re-ranking systems has focused on a two-stage pipeline: \emph{i)} finding a suitable stochastic re-ranking policy, and \emph{ii)} sampling rankings from such a policy, see Fig.~\ref{fig:problem}. 
The pipeline works as follows:
\begin{itemize}[
    leftmargin=0.5cm,
    align=right
]
\item \textbf{Get a re-ranking policy:} We find an appropriate stochastic re-ranking policy by solving a constrained optimization problem for each query.
\item \textbf{Sampling:} We store all possible rankings from every policy in a database. Then, during deployment, we randomly draw a ranking from the pool of stored rankings for each issued query. 
\end{itemize}
This pipeline is both data and memory inefficient. First, it repeats the same optimization procedure several times without generalizing to new queries. As a result, it neither takes advantage of the data nor the task's repetitive nature. In contrast, modern Machine learning systems tackle both issues. Second, it needs to store the set of all possible rankings from a policy for each query, again, without generalization to queries unseen during training.

This paper proposes a learning approach to approximate all stochastic re-ranking policies in a dataset via a shared model across queries capable of generalizing to unseen ones. Our framework relies on meta-learning for learning-to-predict solutions to optimization problems, i.e., finding suitable stochastic re-ranking policies. In particular, we introduce the Constrained Meta-Optimal Transport (CoMOT) method for learning to predict re-ranking policies. This method extends the recently proposed Meta-Optimal Transport~\cite{amos2022meta} to satisfy ranking constraints. CoMOT reduces memory storage by storing a single neural-based model. Additionally, it generalizes to unseen queries during inference.
However, CoMOT only approximates policies and needs an additional step to draw rankings. Therefore, we propose Gumbel-Matching (GumMS), a method for online sampling rankings from stochastic policies, thus, reducing memory storage and generalizing to unseen queries.

\paragraph{Our contributions} 
We summarize our contributions as follows:

\begin{enumerate}[
label={\textbf{(\alph*)}},
leftmargin=0.5cm,
align=right
]
\item Drawing ideas from~\cite{amos2022meta}, we introduce CoMOT, a meta-optimal transport-based approach to learning to predict fair stochastic re-ranking policies. To our knowledge, this paper is the first to use meta-optimal transport to approximate re-ranking policies.

\item Inspired by~\cite{bruch2020stochastic,mena2018learning}, we propose GumMS, an online approach to draw rankings randomly from any doubly-stochastic-based ranking policy.

\item On two TREC datasets, we study CoMOT's ability to approximate and generalize stochastic re-ranking policies. Also, we investigate GumMS's finite-sample capability to approximate doubly-stochastic-based ranking policy in expectation.

\end{enumerate}

\section{Related work}
We refer to~\cite{ekstrand2022fairness} for a detailed overview of fair ranking. 
\subsection{Re-ranking with stochastic policies}
%Randomization arises in many information retrieval (IR) systems. 
Pandey et al.~\cite{pandey2005shuffling} first proposed randomized ranking motivated by click exploration. Furthermore, randomness has been shown to improve diversity in ranking~\cite{radlinski2008learning}, and in search results~\cite{diaz2020evaluating,gao2020toward} 

Specifically, stochastic ranking policies enforce performance requirements in expectation, e.g., maximizing expected utility. These policies are often encoded as Doubly-Stochastic (DS) matrices, i.e., non-negative matrices whose rows and columns sum to one. Finding such policies involves solving a constrained optimization problem.

A stochastic re-ranking system is a stochastic ranking policy on top of a pre-trained ranker. Thus, a re-ranking system acts as a post hoc step that receives a pre-sorted list and outputs a new sorting that satisfies prespecified properties that the original ranker did not, e.g., fairness or calibration.

\subsubsection{Fair stochastic policies}
We can include more constraints to the optimization problem
such that the stochastic ranking system displays a desired fair behavior.
There are mainly two axes of research in fair rankers: \emph{i)} to define fairness constraints and \emph{ii)} to devise approaches to solve the optimization problem, i.e., using general-purpose or custom solvers. 

Singh and Joachims~\cite{singh2018fairness} propose a stochastic ranking policy that exposes different sub-groups equally in expectation, i.e., fairness of exposure (FOE). Biega et al .~\cite{biega2018equity} also rely on exposure to define an alternative notion of fairness for rankings. Later works combine FOE with additional constraints. Wang and Joachims~\cite{wang2021user} add diversity restrictions, and Singh et al.~\cite{singh2021fairness} include uncertainty. 

Do and Usunier~\cite{do2022optimizing} enforce optimizing generalized Gini indices for fair rankings and propose a custom solver based on conditional gradient descent. \cite{heuss2022fairness, sarvi2022understanding} focus on reducing the effect of outliers in fair stochastic re-ranking policies. Other works handle multisided fairness criteria~\cite{su2022optimizing,wu2022joint} and learning fair stochastic policies via the policy-gradient approach~\cite{yadav2021policy, singh2019policy}.

All methods above must repeatedly solve a constrained optimization problem to build a stochastic ranking policy per query.
Consequently, these methods do not deal with the task's redundancy nor leverage the data to warm-start the optimization problem. Hence, this work takes a learning-to-optimize perspective to leverage the data and reduce redundancy.

\subsection{Sampling rankings from stochastic policies}
Many DS-based stochastic ranking methods rely on the Birkhoff von Neumann decomposition (BvND) to sample rankings. The BvND represents a DS matrix as a convex combination of permutations/rankings, where the number of such permutations can be quadratic in the number of documents/items. Kletti et al.~\cite{kletti2022introducing} present a decomposition that ensures as many permutations as the number of documents. However, this method is specific to the Dynamic Bayesian Network exposure model~\cite{chapelle2009dynamic}.

\cite{bruch2020stochastic,wu2022joint,oosterhuis2021computationally} draw single permutations from a deterministic ranker by perturbing its scores using the Concrete distribution~\cite{maddison2016concrete}. In particular, Bruch et al.~\cite{bruch2020stochastic} show that it is possible to train rankers that optimize expected metrics computed over multiple rankings sampled based on estimated relevance.

This work uses the Gumbel-matching  distribution~\cite{mena2018learning}
 to sample rankings from any DS-based policy. 

\subsubsection{Deterministic policies}
Zehlike et al .~\cite{zehlike2017fa} present a re-ranking algorithm that guarantees a percentage of items from the protected groups in every prefix of the top-k. Celis et al .~\cite{celis2017ranking} frame the re-ranking problem as a constrained optimization problem. 

Other works include fairness-based objectives in the learning process. In particular, these methods optimize a learning loss based on an ideal stochastic ranking policy. Zehlike and Castillo~\cite{zehlike2020reducing} propose DELTR, a modification of ListNet~\cite{cao2007learning} that includes the FOE constraint. However, its original version employs a deterministic ranker without post hoc randomization. Kotary et al.~\cite{kotary2022end} introduce a Teacher-Student approach. It minimizes the regret of the ranker against the oracle solution, e.g., the FOE problem. This method requires solving a linear program for each query and training epoch.

This work aims at building a stochastic re-ranking pipeline that returns, for each function call, a ranking that is fair in expectation.

\section{Preliminaries}
\paragraph{Notation}
%We first introduce the necessary notation.
%
Bold lower-case letters denote column vectors, e.g.,  
$\rvx \in \sR^{n}$, $\vone_n$ is the one-vector of size $n$, bold capital letters are matrices, e.g., 
$\rmX \in \sR^{n\times n}$, and $[n]$ denotes $\{1, \ldots, n\}$.  
Let $\rmC \in \sR_{+}^{n\times n}$ be the cost matrix for $\rvx,\rvy \in \sR^{n}$. 
We compute each entry $\rmC_{ij} = c(\rvx_i, \, \rvy_j)$ with the cost function $c: \sR \times \sR \rightarrow \sR_{+}$ for all $(i, j) \in [n]^2$. 
%Let  $\langle \cdot, \cdot \rangle\frob$ be the matrix inner product, $\left\langle \rmP, \, \rmC \right\rangle\frob = \sum_{(i,j) \in [n]^2} \rmC_{ij} \, \rmP_{ij}$ for $\rmC$ and $\rmP \in \sR^{n\times n}$.

Let $\gP_n \coloneqq \left\{\rmT \in \{0, 1\}^{n \times n}:\,  \rmT\,  \vone_n = \vone_n,\,  \vone_n\transpose \, \rmT =  \vone_n\transpose\right\}$ be the set of permutations of $n$ dimensions.

Let $\gD\gS_n \coloneqq \left\{\rmP \in \sR_{+}^{n \times n}: \, \rmP \, \vone_n = \vone_n, \, \rmP\transpose \, \vone_n = \vone_n  \right\}$ be the set of doubly-stochastic (DS) matrices of $n$ dimensions.

We denote $\gQ = \left\{(q,\, \rmX^q)\right\}_{q=1}^{N}$ the set of all query and document-features, $\rmX \in \sR^{n_q\times p}$ is a matrix of $p$ features for $n_q$ documents in query $q$, and $N$ is the number of queries in the dataset.

\subsection{Optimal transport and ranking}
\subsubsection{Kantorovich formulation}
Discrete Optimal Transport (OT) finds an optimal transport matrix given the displacement cost between two random vectors while preserving their marginals or mass.  
Formally, OT ensures an assignment matrix, i.e., ranking, $\rmP \in \gP_n$ that minimizes the cost $\rmC \in \sR^{n \times n}$, 
\begin{equation}
\label{eq:linear_program_without_constrains}
\rmP^{\ast}\left(\rmC\right) \in
\argmin_{
    \rmP \in \gP_n
} 
\sum_{(i,j) \in [n]^2} \rmC_{ij} \, \rmP_{ij}.
\end{equation}
Eq.~\ref{eq:linear_program_without_constrains} encloses pairwise additive metrics~\cite{agarwal2019general}.  For instance, the Discounted Cumulative Gain (DCG) corresponds to $\rmC = \rvu \, \rvv\transpose$, 
where $\rvu \in \sR^n$ is the vector of relevances for $n$ items, and $\rvv_j = -1 / \log_2(1 + j)$ is the discount for each rank position $j \in [n]$. Thus, we can express rankings as solutions to discrete OT problems~\cite{peyre2019computational}, making OT appealing in IR.
Furthermore, we can solve Eq.~\ref{eq:linear_program_without_constrains} with the Hungarian method~\cite{kuhn1955hungarian} with an $\gO(n^3)$ worst-case run-time complexity.

We set the OT cost as the negative of the utility~\cite{singh2018fairness}\footnote{Utility is the negative of the optimal transport cost.}. Thus, we assume the correct signs to perform minimization.

\subsubsection{Entropic OT~\cite{cuturi2013sinkhorn}} 
It adds an entropic penalization term to the original OT problem, Eq.~\ref{eq:linear_program_without_constrains}. This modification smooths out the OT matrix $\rmP$, relaxing the solution set from permutations $\gP_n$ to DS matrices $\gD\gS_{n}$. Entropic OT has the following form:
\begin{equation}
    \rmP^\ast(\rmC, \epsilon) \in  \argmin_{\rmP \in \gD\gS_{n}} 
    %{\left\langle \rmP, \, \rmC \right\rangle\frob} 
    \sum_{(i,j) \in [n]^2} \rmC_{ij} \, \rmP_{ij}
    - \epsilon\, {H(\rmP)},
\end{equation}
where $H(\rmP) \coloneqq - \sum_{i, j} \rmP_{ij} \left(\log \left(\rmP_{ij}\right) - 1 \right)$ is the entropic regularization term, and $\epsilon > 0$ is the regularization value that controls the smoothness of the solution. 
Accordingly, a DS-based re-ranking policy $\rmP \in \gD\gS_n$ encodes the probability of swapping an item from position $i$ to position $j$, for all $(i, j) \in [n]^2$. $\rmP$ is also called the marginal rank probability (MRP) matrix~\cite{heuss2022fairness}. Furthermore, entropic OT can be solved using the Sinkhorn matrix scaling~\cite{sinkhorn1967concerning}, Algorithm~\ref{alg:sinkhorn},  which is end-to-end differentiable.

Cuturi et al.~\cite{cuturi2019differentiable} defined a differentiable ranking algorithm based on entropic OT. It replaces the set of permutation matrices with doubly stochastic (DS) matrices. Nonetheless, we can trace the idea of relaxing the permutation ranking into a DS matrix back to~\cite{adams2011ranking}.

As our objective is re-ranking, we restrict OT-related definitions to uniform marginals, i.e., the transport matrix is DS up to the number of documents in the query.

\subsection{Probabilistic rankings and fairness}
\subsubsection{Stochastic ranking policies}
Rankings are combinatorial objects. Therefore, finding a unique solution that satisfies constraints, such as ensuring fair exposure of various groups, may not even exist. Thus, we lean on achieving constraints in expectation. There are many ways to satisfy restrictions in expectations, such as amortizing through time, e.g., by displaying certain groups during the weekends. We can also sample from an appropriate distribution over rankings, i.e., a stochastic re-ranking policy. In this work, we focus on the latter, as we can represent a stochastic policy as a DS matrix, $\rmP \in \gD\gS_n$.

The Birkhoff–von Neumann decomposition (BvND)~\cite{birkhoff1940lattice} serves as the primary method to sample from DS matrices. The BvND states that any DS matrix $\rmP \in \gD\gS_n$ can be decomposed as a convex combination of permutation matrices~\cite{birkhoff1946tres,dufosse2016notes}, which can be represented as $\rmP = \sum_{i=1}^{k} \alpha_i \rmT^{i}$. Here, $\sum_{i=1}^{k} \alpha_i = 1$,  $0 \leq  \alpha_i \leq 1$, and $\rmT^{i} \in \gP_n$ for $i \in [k]$.
Therefore, we can sample from a DS-based policy by selecting a permutation matrix $\rmT^{i}$ from the BvND with a probability proportional to its coefficient $\alpha_i$~\cite{singh2018fairness}.
There exists a BvND with $k < (n - 1)^2 + 1$~\cite{dufosse2018further}, leading to the assumption that $k=\gO(n^2)$. However, finding the minimum $k$ is an NP-hard problem.

\subsubsection{Constrained stochastic ranking policies}
A stochastic ranking policy $\rmP \in \gD\gS_n$ ensures a minimum transport cost in expectation. 
Nevertheless, we can impose additional constraints by including more of them in the optimization problem, Eq.~\ref{eq:linear_program_without_constrains}, as follows:
\begin{equation}
\label{eq:linear_program_with_constrains}
\rmP^{\ast} \in 
\argmin_{
    \rmP \in \gD\gS_n
} 
\sum_{(i,j) \in [n]^2} \rmC_{ij} \, \rmP_{ij}
\quad \text{s.t. } \Omega(\rmP) \leq \rho,
\end{equation}
where $\Omega: \gD\gS_n \rightarrow \sR_{+}$ is the constraint, e.g., fairness of exposure, and $\rho$ denotes penalty level. We can solve Eq.~\ref{eq:linear_program_with_constrains} with cubic complexity using Linear Programming whenever $\Omega(\cdot)$ is linear in $\rmP$.  
Singh and Joachims~\cite{singh2018fairness} proposed several linear constraints $\Omega$.

\section{Learning to re-rank with C\lowercase{o}MOT}
In the previous section, we presented the link between OT and ranking. 
We noted that current re-ranking pipelines do not leverage information across the data, as it stores many fixed-permutation matrices for each issued query. Instead, we propose to learn a single predictive-approximative re-ranking policy model for all queries, taking advantage of data redundancies backed up with an online sampling approach. In particular, given the OT-ranking relation and the necessity to solve multiple OT problems, we frame the stochastic re-ranking policy learning as a \textbf{Co}nstrained \textbf{M}eta-\textbf{O}ptimal \textbf{T}ransport problem, CoMOT. CoMOT relies on Meta-Optimal transport (MOT), which aims at learning the shared structure and correlations between multiple OT solutions.

\subsection{FOE: Fairness of exposure}
Singh and Joachims~\cite{singh2018fairness} introduced various notions of fairness of exposure in ranking and used the average group exposure as a foundation.
Let $\gG_{l}$ be the list of documents belonging to group $l$, e.g., $l$ can represent \emph{gender}. Then, given a stochastic re-ranking policy $\rmP \in \gD\gS_n$, the average exposure for $\gG_{l}$ is:
\begin{equation}
\label{eq:exposure}
\Exposure(\gG_{l}|\,  \rmP) = \frac{1}{|\gG_{l}|}\sum_{i \in \gG_{l}} \sum_{j=1}^{n} \rmP_{i, j} \rvv_j,
\end{equation}
where $\rvv_j$ is the discount factor that encodes the importance of rank position $j$ for all $j \in [n]$, e.g., logarithmic discount.

This work focuses on FOE as a fairness constraint $\Omega(\cdot)$. For two groups, $l$ and $l^\prime$, FOE enforces their average group exposure to be identical, as follows
\begin{equation}
    \label{eq:demographic_parity_equality}
    \Omega(\rmP) = g\left(\Exposure(\gG_l|\, \rmP) - \Exposure(\gG_{l^\prime}|\, \rmP) \right),
\end{equation}
where $g(\cdot)$ is preferably a (sub)differentiable function. 
We focus on two groups, i.e., \emph{protected} and \emph{non-protected}, and use $g(x) = |x|$.

\subsection{MOT: Meta-Optimal Transport}
\subsubsection{Setting up MOT}
MOT~\cite{amos2022meta} uses amortized optimization~\cite{amos2022tutorial} to maximize the dual objective of the entropic OT problem. Optimizing the dual objective involves finding two potential functions. Mainly, MOT simplifies the problem by linking these potential functions using the optimal solution's characterization. 
Therefore, we only fit a neural network that approximates one potential function.

As presented in~\cite{peyre2019computational}, the dual solution of the entropic OT is
\begin{equation}
    \label{eq:entropic_dual}
    \rvf^\ast, \rvg^\ast \in 
    \argmax_{\rvf\in \sR^{n}, \, \rvg\in \sR^{n}} 
    \langle \rvf, \rva \rangle + \langle \rvg, \rvb \rangle - 
    \epsilon\, \langle 
    \exp\left(\rvf/\epsilon\right),
    \rmK \exp\left(\rvg/\epsilon\right) 
    \rangle,
\end{equation}
where $\rmK_{ij} = \exp\left(-\rmC_{ij}/\epsilon\right)$ is the Gibbs kernel and the dual variables or potentials $\rvf\in \sR^{n}$ and $\rvg\in \sR^{n}$ are
associated, respectively, with the marginal constraints $ \rmP \, \vone_n = \vone_n$ and $\rmP\transpose \, \vone_n = \vone_n$.
The optimal duals depend on the problems, $f^\ast(\rmC, \epsilon)$, but we omit this dependence for notational simplicity.

We can use the first-order optimality condition of Eq.~\ref{eq:entropic_dual} to map between the optimal dual potentials,
\begin{equation}
\label{eq:dual_potential}
    g(\rvf; \,\rmC, \epsilon) \coloneqq \epsilon\, \log \vone_n - \epsilon\, \log \left(\rmK\transpose \, \exp\left(\rvf/ \epsilon\right) \right).
\end{equation}

Additionally, we can obtain the optimal coupling $\rmP$ from the optimal duals $\rvf^\ast$ and $\rvg^\ast$, as follows
\begin{equation}
    \label{eq:primal_from_dual}
    \rmP_{ij}^\ast(\rvf;\, \rmC, \epsilon) \coloneqq \exp(\rvf_i^\ast / \epsilon)\, \rmK_{ij} \, \exp(\rvg_j^\ast / \epsilon), \quad \forall (i, j) \in [n]^2,
\end{equation}
where $\rvg = g(\rvf; \, \rmC, \epsilon)$ comes from evaluating Eq.~\ref{eq:dual_potential}.

\subsubsection{MOT learning procedure}
MOT has the following learning objective: 
\begin{equation}
    \min_{\theta} \frac{1}{N}\sum_{q=1}^{N} J\left(\rvf;\, \rmC^{q}, \epsilon \right), \quad \text{where } \rvf = f_\theta(\rmC^q),
\end{equation}
where $f_{\theta}(\cdot)$ is the potential model with parameters $\theta$, e.g., a neural network, 
$\rmC^q$ denotes the transport cost for query $q$, $N$ is the number of queries in the dataset, 
and $-J(\rvf;\, \rmC, \epsilon)$ is the amortized objective:
\begin{equation}
    \label{eq:mot_loss}
    J(\rvf; \, \rmC, \epsilon) = \langle \rvf + \rvg, \, \vone_{n_q}  \rangle
    - \epsilon\, \left\langle 
    \exp\left(\rvf\, /\epsilon\right),
    \rmK \exp\left(\rvg\, /\epsilon\right) 
    \right\rangle.
\end{equation}
At optimality, the pair of potentials are related via Eq.~\ref{eq:dual_potential}. It is thus sufficient to find only one of them, e.g., $f_{\theta}$. Besides, the MOT procedure does not require a ground truth solution or a teacher, as it uses amortization as a learning paradigm.

\subsection{Sampling permutations from OT maps\label{sec:gumbel}}
Gumbel-Matching~\cite{mena2018learning} samples multiple rankings by applying a perturbation to the transport cost, a.k.a., Gumbel-trick~\cite{jang2016categorical,gane2014learning}.  
Particularly, a random permutation $\rmP \in \gP_{n}$ follows the Gumbel-Matching distribution with parameter $\rmC$, denoted $\rmP \sim \gG\gM(\rmC)$,  if  $\rmP$ solves Eq.~\ref{eq:linear_program_without_constrains} with cost matrix $\tilde{\rmC} = \left(\rmC + \sigma\, \rmN \right) / \tau$, where $\rmN \in \sR^{n \times n}$ is a matrix of standard i.i.d. Gumbel noise, $\sigma > 0$ is the noise level, and $\tau > 0$ is the temperature.  
Consequently, sampling from $\gG\gM(\rmC)$ boils down to solving Eq.~\ref{eq:linear_program_without_constrains}, which can be solved with the Hungarian method\cite{kuhn1955hungarian,peyre2019computational} \footnote{The Hungarian method solves Eq.~\ref{eq:linear_program_without_constrains} only for marginals $\vone_n /n$ instead of $\vone$. Thus, we rescale the marginals and the resulting transport mapping accordingly.} with an $\gO(n^3)$ worst-case run-time complexity.

\subsection{Proposed solution}
\begin{figure*}[!ht]
	\small
	\centering
	{\includegraphics[width=1\linewidth, trim={0 0mm 0 0mm}, clip]{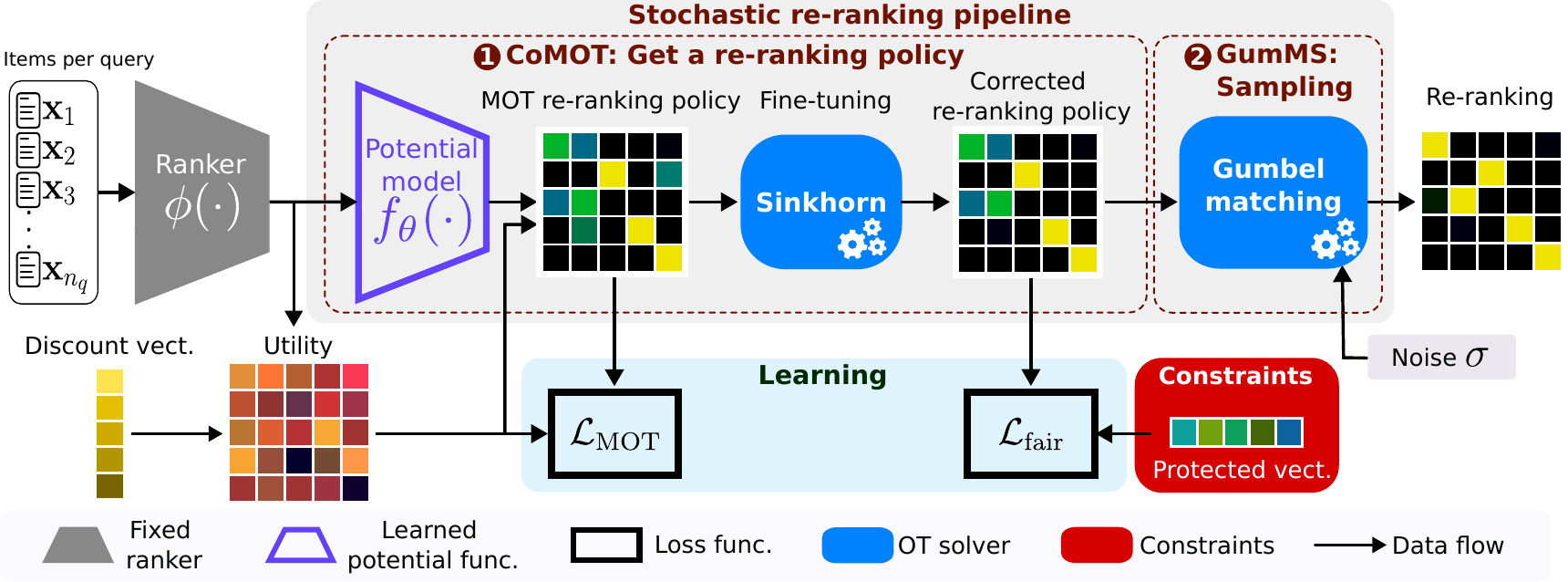}}
	\caption{
The proposed framework. A pre-trained ranker assigns a score to each item in the list for each query. Then, we use meta-optimal transport to learn a potential function shared across queries. On average, this potential function optimizes the linear assignment of scores to discount positions for all queries. To impose the fairness constraint, we explicitly compute the approximated re-ranking policy, finetune it using the Sinkhorn algorithm, and apply it to calculate the fairness loss. Finally, we sample a ranking from the re-ranking policy via Gumbel-Matching for each function call during deployment.
}
	\Description{
We use constrained meta-optimal transport to learn a potential model shared across queries to approximate query-based ranking policies.  
Also, we use Gumbel-Matching to sample rankings from probabilistic ranking policies.
    }
	\label{fig:solution}    
\end{figure*}

We propose a strategy to reduce memory storage and leverage data in constrained re-ranking systems. Our proposed method receives a list of scaled ranking scores for every query. It returns a ranking/permutation that satisfies a pre-specified ranking constraint in expectation, e.g., a FOE fairness constraint. This method requires only storing the parameters of a potential model, e.g., a neural network, instead of $\gO(N \, n^3)$ in current approaches. 
Fig.~\ref{fig:solution} depicts our system, which has two stages: \emph{i)} offline learning-to-predict a fair re-ranking policy, and \emph{ii)} online sampling rankings from such a policy. The first stage needs training a potential model $f_{\theta}$, which is shared across queries. The second stage requires solving a linear assigning problem for each call during deployment.

\subsubsection{Predicting fair re-ranking policies with CoMOT}
We compute the OT cost for each query $q$, $\rmC^{q} \in\sR^{n_q \times n_q}$. This cost is the ``price'' of assigning a ranking score to a discounted position. Then, we evaluate the potential model $\rvf = f_{\theta}(\rmC)$, compute its MOT loss, $\gL_{\MOT} = -J(\rvf; \rmC^q, \epsilon)$, and get its associated ranking policy, $\rmP^{\MOT}$. However, the predicted ranking policy $\rmP^{\MOT}$ is usually not DS; one of its marginals does not add up to one. 

Since the MOT policy is not DS, enforcing any constraint directly into the dual form may hamper the performance of the re-ranking policy, as numerically, the algorithm might deceive the potential model into believing its ranking policy satisfies the re-ranking constraint as it minimizes the amortized cost. However, it fails once we project onto the DS manifold. As a result, we need to finetune the MOT solution before enforcing any constraint. We use the Sinkhorn algorithm to obtain DS matrices, Algorithm~\ref{alg:sinkhorn}. This algorithm has $\gO(n^2)$  computation complexity as it iteratively rescales rows and columns of a matrix $\rmP \in \sR^{n \times n}$ for a fixed number of iterations $L$ so that each dimension sums to 1.
\begin{algorithm}[!ht]
\caption{Sinkhorn algorithm for DS matrices~\cite{sinkhorn1967concerning}}
\label{alg:sinkhorn}
\begin{algorithmic}[1]
    \REQUIRE{A square matrix $\rmP \in \sR^{n\times n}$,  entropic parameter $\epsilon$, and number of iterations $L$.}
    %\ENSURE{A fair ranker $\phi^{\ast}$}    
    %\Statex{\textit{Initialize Ranker}}
    \STATE{$\rmA \gets \rmP / \epsilon$}
    \FOR{$l \in [L]$}
        \FOR{$(i,j) \in [n]^2$}
            \STATE{$\rmA_{i j} \gets \rmA_{i j} - \log \sum\limits_{j^\prime \in [n]} \exp \left(\rmA_{i j^\prime}\right)$}
            \STATE{$\rmA_{i j} \gets \rmA_{i j} - \log \sum\limits_{i^\prime \in [n]} \exp \left(\rmA_{i^\prime j}\right)$}
        \ENDFOR{}
    \ENDFOR{}
    \RETURN{$\exp\left(\rmA\right)$}
\end{algorithmic}
\end{algorithm}

Once we get the MOT ranking policy $\rmP^{\MOT}$, we impose fairness constraints by minimizing the fairness loss, $\gL_{\fair}$ (see Eq.~\ref{eq:demographic_parity_equality}). Hence, the overall loss to minimize is $\gL_{\MOT} + \lambda\, \gL_{\fair}$, where $\lambda > 0$ is the fairness regularization that controls the influence of the fairness constraint on the total loss.

Algorithm~\ref{alg:proposed_algorithm} summarizes the CoMOT learning scheme, which finds a well-suited potential model $f_{\theta}$ that minimizes the overall population loss. First, the algorithm receives ranking scores from a pre-trained ranker for each query and standardizes them by min-max scaling. Then, it computes the transport cost $\rmC^q$ and predicts the values of the OT potential using the potential model $f_{\theta}$. With this cost and potential, it gets an approximated transport map and rescales it using the Sinkhorn algorithm. 
Next, it computes the fairness loss using the protected items and the transport map. Finally, it combines the MOT and fairness loss and backpropagates to update the parameters $\theta$ of the potential model. We repeat the previous steps for all queries and several training epochs.  
\begin{algorithm}[!ht]
\caption{CoMOT: The optimization scheme for Learning to re-rank with Constrained Meta-Optimal Transport}
\label{alg:proposed_algorithm}
\begin{algorithmic}[1]
    \REQUIRE{
        ranker $\phi$,
        entropic regularization $\epsilon$, 
        fairness regularization $\lambda$, 
        and number of Sinkhorn iterations $L$
    }
    \STATE{Initialize Potential model, $f_{\theta}$}  
    \REPEAT{}
        \FOR{$(q,\, \rmX^q) \in \gQ$}
            \STATE{$\rvu \gets \phi(\rmX^q), \quad \rvu \in \sR^{n_q}$}
            \COMMENT{Document scoring}
            \STATE{$\tilde{\rvu} \gets \frac{\rvu - \min(\rvu)}{\max(\rvu) - \min(\rvu)}$}
            \COMMENT{ Min-Max scaling}
            \STATE{$\rmC^q \gets \tilde{\rvu}\, \rvv\transpose, \quad \rmC^q \in \sR_{+}^{n_q \times n_q}$}
            \COMMENT{Transport cost matrix}
            \STATE{ $\rvf \gets f_{\theta}(\tilde{\rvu}), \quad \rvf \in \sR^{n_q}$}
            \COMMENT{Predict potential values}

            \STATE{$\rmP_{ij}^{\MOT} \gets \rmP^{\ast}_{ij}\left(\rvf;\,\rmC^q, \epsilon\right), \quad \forall (i, j) \in [n]^2$}
            \COMMENT{Get policy, Eq.~\ref{eq:primal_from_dual}}
            \STATE{$\tilde{\rmP}^{\MOT} \gets \textsf{Sinkhorn}(\rmP^{\MOT}, \epsilon, L)$}
            \COMMENT{Fine-tuning, Eg.~\ref{alg:sinkhorn}}
            \STATE{$\gL_{\fair} \gets \Omega(\tilde{\rmP}^{\MOT})$}
            \COMMENT{Fairness loss, Eq.~\ref{eq:demographic_parity_equality}}
            \STATE{$\gL_{\MOT} \gets  -J(\rvf;\, \rmC^q, \epsilon)$}    
            \COMMENT{$\MOT$ loss, Eq.~\ref{eq:mot_loss}}            
            \STATE{$\gL \gets \gL_{\MOT} + \lambda\, \gL_{\fair}$}
            \COMMENT{Compute total loss}
            \STATE{$\grad_{\theta} \gets \AutoDiff_{\theta}\left(\gL\right)$}
            \COMMENT{Compute gradient}
            \STATE{$\theta \gets \theta - \alpha\cdot \AdamOperator(\grad_{\theta})$}
            \COMMENT{Update potential model}
       \ENDFOR{}
    \UNTIL{Convergence}
    \RETURN{Potential model ${f}_{\theta}$ that produces $\rmP^{\CoMOT}$} 
\end{algorithmic}
\end{algorithm}

During deployment, the CoMOT algorithm has $\gO(n^2)$ computational complexity, similar to self-attention models~\cite{vaswani2017attention}. 

We note that the transport map $\rmP^{\MOT}$ has not reached a fixed point for midway epochs in Algorithm~\ref{alg:proposed_algorithm}. As a result, the Eq.~\ref{eq:primal_from_dual} does not hold for intermediate training epochs. Despite this, the final map minimizes the fairness loss in practice as we repeat parameter updates several times. However, the fairness regularization parameter $\lambda$ must usually be large. Similar behavior has been reported in bi-level optimization approaches, where methods are often used online when the inner parameters are far from convergence~\cite{vicol2021implicit}.

\subsubsection{GumMS: Sampling rankings from a ranking policy}
As presented in Sec.~\ref{sec:gumbel}, sampling from a Gumbel-Matching (GM) distribution requires modifying the cost matrix. However,  CoMOT produces smooth policies, and small cost perturbations may lead to unperceived modifications to the policy. Instead, we propose to use  $\rmC = 1 - \rmP$ as the transport cost parameter for the GM distribution. Given that the policy $\rmP_{ij}$ encodes the probability of re-ranking positions $i$ and $j$, this cost is still positive, and higher values of $\rmP$ correspond to lower costs and hence are more likely to match.

The Algorithm~\ref{alg:proposed_sampling} depicts our sampling strategy to draw rankings $\rmT \in \gP_n$ from a stochastic ranking policy $\rmP \in \gD\gS_n$. We sample $\rmT$ from a GM distribution with parameters $1 - \rmP$, $\rmT \sim \gG\gM\left(1 - \rmP\right)$.  
\begin{algorithm}[!ht]
\caption{GumMS: Sampling rankings from a DS-based policy}
\label{alg:proposed_sampling}
\begin{algorithmic}[1]
    \REQUIRE{
        Policy $\rmP \in \gD\gS_{n}$, temperature $\tau$, and noise level $\sigma$.
    }
    \STATE{$\rmN \sim \Gumbel$}
    \STATE{$ \tilde{\rmC} \gets  ((1 - \rmP) + \sigma\, \rmN) / \tau$}
    \COMMENT{Perturbed cost}
    \STATE{$\rmT^\ast \gets  \argmin\limits_{\rmT \in \gP_{n}} \sum\limits_{(i,j) \in [n]^2} \tilde{\rmC}_{ij} \, \rmT_{ij}$}
    \COMMENT{Optimal assignment~\cite{kuhn1955hungarian}}
    %\COMMENT{Solve by Hungarian method~\cite{kuhn1955hungarian}}
    \RETURN{Permutation matrix $\rmT^{\ast} \in \gP_{n}$}
\end{algorithmic}
\end{algorithm}

\section{Experiments}
In this section, we analyze the proposed re-ranking pipeline with respect to the solution to original constrained optimization. Specifically, we will study the following research questions:
\begin{itemize}
\item[\rqone{}:] Does CoMOT provide a good approximation to fair stochastic re-ranking policies?
%MOT relies on a smooth relaxation of the transport problem. This relaxation affects performance.

\item[\rqtwo{}:] What are the computation time, ranking, and fairness performance of the  CoMOT's predicted policy compared to an ideal policy found by solving a constrained optimization problem?

\item[\rqthree{}:] Does the GumMS online sampling ranking approach approximate the original re-ranking policy in expectation?

\end{itemize}

\subsection{Experimental settings}

\subsubsection{Datasets}
We use two academic search datasets in our experiments, 
the TREC2019 and TREC2020 Fair Ranking track\footnote{\url{https://fair-trec.github.io/}}. These datasets come with queries, relevance judgments, and information about the authors and academic articles extracted from the Semantic Scholar Open Corpus\footnote{\url{http://api.semanticscholar.org/corpus/}}. 

Table~\ref{tab:datasets} summarizes the descriptive statistics of the datasets. Our pre-processing step removes queries with only one protected group or less than three documents. The TREC2020 dataset comes with 200 queries for training and 200 for testing; after pre-processing, we end up with 191 and 190 queries in the train and test sets, respectively. The TREC2019 dataset comes with 631 queries for training and 635 for testing, and after pre-processing, we get 346 and 554 queries in the train and test sets, respectively. 
As input to the learning-to-rank (LTR) model, we use the same data as OMIT\footnote{\url{https://github.com/arezooSarvi/OMIT_Fair_ranking}} with 25 features based on term frequencies, BM25~\cite{robertson2009probabilistic}, and language models.
\begin{table}[!ht]
    %\small
    \footnotesize
    \centering
    \caption{Descriptive statistics of the original and pre-processed TREC 2019 and 2020 Fair Ranking track datasets.}
    \label{tab:datasets}
    \begin{tabular}{l@{\hspace{6mm}}rrrrr}\\
    \toprule
    
    \multirow{2}{*}{Dataset:}
    & \multicolumn{2}{c}{\textsf{TREC2019}} & \multicolumn{2}{c}{\textsf{TREC2020}} \\
    \cmidrule(lr){2-3} \cmidrule(lr){4-5}
    & Train & Test & Train & Test \\
    \midrule
    \textsf{Original} &&&&\\ \quad Avg. \#\,items per query & 4.1 & 6.8& 23.5 & 23.4\\
\quad Queries & 631 & 635& 200 & 200\\
\quad Avg. \#\,rel. items per query & 2 & 3.3& 3.7 & 3.4\\
 \midrule \textsf{Pre-processed} &&&&\\\quad Avg. \#\,items per query & 5.1 & 6.9& 23.8 & 23.9\\
\quad Queries & 346 & 554& 191 & 190\\
\quad Avg. \#\,rel. items per query & 2.4 & 3.3& 3.8 & 3.4\\

\bottomrule
\end{tabular}
\end{table}

\subsubsection{Baselines}
Our re-ranking policy is model and loss agnostic. Therefore, we used several combinations of ranking architectures and losses as the backbone to score documents. First, we employed a Linear ranker with various training losses, SVMRank~\cite{joachims2002optimizing}, RankNet~\cite{cao2007learning}, SVMRank DCG~\cite{agarwal2019general}, ListNet~\cite{cao2007learning}, and
$\lambda$Loss nDCG~\cite{wang2018lambdaloss}. Then, we used the ListNet loss to fit several ranking architectures: All-Rank~\cite{pobrotyn2020context},  DASALC~\cite{qin2020neural}, ATTNDIN~\cite{pasumarthi2020permutation}, Linear, and Multi-Layer Perceptron (MLP).

We compare CoMOT against the following baselines:
\begin{itemize}
    \item \textbf{Inital ranking}:  A traditional LTR model. We build on top of the output scores of this model.
    \item \textbf{CVXFOE}~\cite{singh2018fairness}: We rely on the CVX~\cite{diamond2016cvxpy} solver to build a stochastic re-ranking policy by solving Eq.~\ref{eq:linear_program_with_constrains} with Fairness of exposure (FOE) constraints, Eq.~\ref{eq:demographic_parity_equality}, using as input the scores of the initial ranker. 
\end{itemize}

\subsubsection{Evaluation metrics}
We assess the absolute value of the Fairness of Exposure (FOE-abs) as our fairness constraint and use the normalized discounted cumulative gain (nDCG) as a performance indicator. In particular, we use nDCG at five and ten.

To assess the quality of GumMS, we compute the point-wise squared error between the CoMOT policy $\rmP^{\CoMOT}$ and the average of $k$ random samples from GumMS, $\tilde{\rmP}^k = \frac{1}{k}\sum_{i=1}^{k} \rmT^{i}$, where $\rmT^{i} \sim \gG\gM( 1 - \rmP^{\CoMOT})$. We measure the approximation error as follows: 
\begin{equation}
    \label{eq:gumms_evaluation}
    \begin{split}
        \| \rmP^{\CoMOT} - \tilde{\rmP}^k \|^{2}_{\textsf{F}} &= \sum\limits_{(i, j) \in [n]^2} (\rmP_{ij}^{\CoMOT} - \tilde{\rmP}_{ij}^k)^2.
    \end{split}
\end{equation}

\subsubsection{Implementation details.}
We use Pytorch~\cite{NEURIPS2019_9015} to implement our proposed algorithms and rely on PytorchLTR~\cite{jagerman2020accelerated} to handle learning-to-rank (LTR) data. We employ Hydra~\cite{Yadan2019Hydra} to configure and store our experimental settings. We train all models for 30 training epochs and use AdamW~\cite{loshchilov2018decoupled} with a fixed learning rate of $0.001$ to update the model's parameters. We set the batch size to 12 and one for the LTR stage and re-ranking, respectively.

We use CVXPY~\cite{diamond2016cvxpy} to solve the FOE problem, Eq.~\ref{eq:linear_program_with_constrains} with constraints Eq.~\ref{eq:demographic_parity_equality}.  CoMOT uses $L=10$ Sinkhorn iterations, an entropic regularization of $\epsilon=0.1$, and a fairness regularization parameter of $\lambda=1\times10^5$. 
For GumMS, we set $\sigma = 1 / \sqrt{n_q}$, the temperature to 1, and use the Earth Moving Distance algorithm available in the POT library~\cite{flamary2021pot} to obtain a ranking. %Our code is released at \url{CODE}.

The potential model $f_{\theta}$ is a three-layer MLP with 150 hidden units, layer normalization, and ReLU activation. The MLP ranker is a two-layer MLP with 25 hidden neurons, layer normalization, and ReLU. We use two attention heads, 0.2 dropout, and three encoding layers with hidden dimensions $[64, 128, 256]$ for attention-based rankers, DASALC, All Rank, and ATTNDIN. Also, we clamp the output of the potential model to $\pm 5$ to avoid training instability.

\section{Results}

\subsection{CoMOT policy approximation~(\rqone{})}
In this experiment, we compare the predictive performance on testing data of various combinations of ranking architectures and fairness-imposing methods, namely, original ranking results without any corrections (orig.), and CoMOT.  

\subsubsection{Comparing different ranking losses}
This experiment explores the effect of CoMOT re-ranking policy prediction on standard ranking architectures:  a Linear ranker fitted with different LTR losses, pairwise, listwise, and lambda losses. 

Table~\ref{tab:prediction_linear} shows the prediction performance for a Linear ranker trained with various LTR losses. The CoMOT approach reduces the FOE-abs score and keeps the utility cost in all cases. Nevertheless, we observe different behaviors in each dataset. First, the ranking and fairness scores for TREC 2019 are almost the same for all initial rankers. CoMOT-corrected policies present, on average, a five-fold fairness improvement over the initial ranker with a slight gain in ranking assessment.
On the other hand, we observe a uniformizing effect in TREC 2020 after applying CoMOT re-ranking. Notably, the CoMOT re-ranking policy for the Linear ranker trained with SVMRank DCG and ListNet losses displays around 0.29~nDCG@10 with a similar fairness score. This reduction in nDCG contrasts the initial ranking performance, 0.36 and 0.31~nDCG@10 for SVMRank DCG and ListNet, respectively. The value of these prediction scores hints at the possible price of FOE for this data set, i.e., getting the best utility given a defined fairness costs level. 
\begin{table}[!ht]
    %\small
    \footnotesize
    \centering
    \caption{
Comparing the predictive performance of CoMOT's re-ranking for a Linear ranker with different LTR losses on the TREC 2019 and 2020 Fair Ranking datasets. We compare each CoMOT re-ranking against \emph{orig.} using a two-tailed paired t-test ($p < 0.01$).  Statistical significantly lower and higher compared to \emph{orig.} is denoted by $\triangle$ and $\triangledown$ respectively. Bold indicates the best method per policy group.
    }
    \label{tab:prediction_linear}
    \begin{tabular}{cl@{\hspace{3mm}}rrrrr}\\
    \toprule
    & \multirow{2}{*}{Ranking policy} & \multicolumn{2}{c}{\textsf{nDCG$\uparrow$}} & Fairness$\downarrow$  & Utility $\uparrow$ \\
    \cmidrule(lr){3-4} \cmidrule(lr){5-5}
     & & @5 & @10 & FOE-abs & Neg. OT cost \\
    \midrule
    \parbox[t]{2mm}{\multirow{15}{*}{\rotatebox[origin=c]{90}{\textsf{TREC2019}}}}  & RankNet (orig.) & 0.508 & 0.596 & 0.260 & 1.999\\
 & RankNet + CoMOT  & \textbf{0.512}\rlap{$^{\triangle}$} & \textbf{0.598}\rlap{$^{\triangle}$} & \textbf{0.050}\rlap{$^{\triangledown}$} & 1.995\\
 & \rule{-2pt}{3ex} SVMRank (orig.)  & 0.508 & 0.596 & 0.260 & 1.997\\
 & SVMRank + CoMOT  & \textbf{0.512}\rlap{$^{\triangle}$} & \textbf{0.598}\rlap{$^{\triangle}$} & \textbf{0.063}\rlap{$^{\triangledown}$} & 2.009\\
 & \rule{-2pt}{3ex} SVMRank DCG (orig.)  & 0.509 & 0.597 & 0.258 & 1.970\\
 & SVMRank DCG + CoMOT  & \textbf{0.512}\rlap{$^{\triangle}$} & 0.598 & \textbf{0.046}\rlap{$^{\triangledown}$} & 1.946\\
 & \rule{-2pt}{3ex} ListNet (orig.)  & 0.507 & 0.596 & 0.260 & 2.023\\
 & ListNet + CoMOT  & \textbf{0.512}\rlap{$^{\triangle}$} & \textbf{0.598}\rlap{$^{\triangle}$} & \textbf{0.026}\rlap{$^{\triangledown}$} & 1.971\\
 & \rule{-2pt}{3ex} $\lambda$Loss nDCG (orig.)  & 0.510 & 0.596 & 0.255 & 1.942\\
 & $\lambda$Loss nDCG + CoMOT  & \textbf{0.512}\rlap{$^{\triangle}$} & \textbf{0.598}\rlap{$^{\triangle}$} & \textbf{0.059}\rlap{$^{\triangledown}$} & 1.939\\
\midrule
\parbox[t]{2mm}{\multirow{15}{*}{\rotatebox[origin=c]{90}{\textsf{TREC2020}}}}  & RankNet (orig.) & 0.224 & 0.332 & 0.125 & 7.524\\
 & RankNet + CoMOT  & 0.189\rlap{$^{\triangledown}$} & 0.297\rlap{$^{\triangledown}$} & \textbf{0.045}\rlap{$^{\triangledown}$} & 7.640\\
 & \rule{-2pt}{3ex} SVMRank (orig.)  & 0.225 & 0.334 & 0.126 & 7.549\\
 & SVMRank + CoMOT  & 0.189\rlap{$^{\triangledown}$} & 0.297\rlap{$^{\triangledown}$} & \textbf{0.044}\rlap{$^{\triangledown}$} & 7.668\\
 & \rule{-2pt}{3ex} SVMRank DCG (orig.)  & 0.261 & 0.362 & 0.128 & 8.062\\
 & SVMRank DCG + CoMOT  & 0.190\rlap{$^{\triangledown}$} & 0.299\rlap{$^{\triangledown}$} & \textbf{0.042}\rlap{$^{\triangledown}$} & 8.136\\
 & \rule{-2pt}{3ex} ListNet (orig.)  & 0.206 & 0.312 & 0.122 & 6.181\\
 & ListNet + CoMOT  & 0.188\rlap{$^{\triangledown}$} & 0.296\rlap{$^{\triangledown}$} & \textbf{0.056}\rlap{$^{\triangledown}$} & 6.386\\
 & \rule{-2pt}{3ex} $\lambda$Loss nDCG (orig.)  & 0.221 & 0.322 & 0.122 & 6.531\\
 & $\lambda$Loss nDCG + CoMOT  & 0.188\rlap{$^{\triangledown}$} & 0.297\rlap{$^{\triangledown}$} & \textbf{0.045}\rlap{$^{\triangledown}$} & 6.688\\

    \bottomrule
    \end{tabular}
\end{table}

\subsubsection{Comparing different ranking architectures}
Now, we study various neural-based ranking architectures trained with the ListNet loss. We will keep this configuration for the remaining experiments.

Table.~\ref{tab:prediction} shows the predictive performance of rankers with various fairness-constrained methods: the original ranking and CoMOT re-ranking. CoMOT consistently displays lower fairness costs with similar ranking performance and utility costs. However, these fairness gains are not uniform across pre-trained rankers. In the TREC 2019 dataset, for instance,  All Rank + CoMOT reduces FOE by 14, and Linear + CoMOT has a 2.5 times lower FOE. Other rankers + CoMOT fall between these two extreme cases. For the TREC 2020, CoMOT's fairness improvement is less extreme. Nevertheless, we observe again that CoMOT-based re-ranking policies have a lower nDCG. However, this effect is only significant for the Linear and All Rank rankers.
\begin{table}[!ht]
    %\small
    \footnotesize
    \centering
    \caption{
    The predictive performance of CoMOT's re-ranking for various initial ranking architectures.  
    We compare each CoMOT against \emph{orig.} using a two-tailed paired t-test ($p < 0.01$).
    Statistical significance is denoted the same as Table~\ref{tab:prediction_linear}
    }
    \label{tab:prediction}
    \begin{tabular}{cl@{\hspace{3mm}}rrrrr}\\
    \toprule
    & \multirow{2}{*}{Ranking policy} & \multicolumn{2}{c}{\textsf{nDCG$\uparrow$}} & Fairness$\downarrow$  & Utility $\uparrow$ \\
    \cmidrule(lr){3-4} \cmidrule(lr){5-5}
     & & @5 & @10 & FOE-abs & Neg. OT cost \\
    \midrule
    \parbox[t]{2mm}{\multirow{15}{*}{\rotatebox[origin=c]{90}{\textsf{TREC2019}}}}  & \textsf{Linear} (orig.) & 0.507 & 0.596 & 0.260 & 2.023\\
 & \textsf{Linear} + CoMOT  & \textbf{0.512}\rlap{$^{\triangle}$} & \textbf{0.598}\rlap{$^{\triangle}$} & \textbf{0.039}\rlap{$^{\triangledown}$} & 1.991\\
 & \rule{-2pt}{3ex} \textsf{MLP} (orig.)  & 0.507 & 0.596 & 0.260 & 2.720\\
 & \textsf{MLP} + CoMOT  & 0.512 & 0.598 & \textbf{0.091}\rlap{$^{\triangledown}$} & 2.722\\
 & \rule{-2pt}{3ex} \textsf{All Rank} (orig.)  & 0.512 & 0.599 & 0.259 & 2.599\\
 & \textsf{All Rank} + CoMOT  & 0.512 & 0.598 & \textbf{0.018}\rlap{$^{\triangledown}$} & 2.506\\
 & \rule{-2pt}{3ex} \textsf{AttnDin} (orig.)  & 0.512 & 0.599 & 0.265 & 2.912\\
 & \textsf{AttnDin} + CoMOT  & 0.512 & 0.598 & \textbf{0.101}\rlap{$^{\triangledown}$} & 2.935\\
 & \rule{-2pt}{3ex} \textsf{DASALC} (orig.)  & 0.514 & 0.598 & 0.259 & 2.930\\
 & \textsf{DASALC} + CoMOT  & 0.512 & 0.598 & \textbf{0.104}\rlap{$^{\triangledown}$} & 2.952\\
\midrule
\parbox[t]{2mm}{\multirow{15}{*}{\rotatebox[origin=c]{90}{\textsf{TREC2020}}}}  & \textsf{Linear} (orig.) & 0.206 & 0.312 & 0.122 & 6.181\\
 & \textsf{Linear} + CoMOT  & 0.188\rlap{$^{\triangledown}$} & 0.296\rlap{$^{\triangledown}$} & \textbf{0.056}\rlap{$^{\triangledown}$} & 6.386\\
 & \rule{-2pt}{3ex} \textsf{MLP} (orig.)  & 0.219 & 0.322 & 0.120 & 5.310\\
 & \textsf{MLP} + CoMOT  & 0.188 & 0.297 & \textbf{0.050}\rlap{$^{\triangledown}$} & 5.525\\
 & \rule{-2pt}{3ex} \textsf{All Rank} (orig.)  & 0.220 & 0.319 & 0.130 & 5.383\\
 & \textsf{All Rank} + CoMOT  & 0.188\rlap{$^{\triangledown}$} & 0.298\rlap{$^{\triangledown}$} & \textbf{0.022}\rlap{$^{\triangledown}$} & 5.399\\
 & \rule{-2pt}{3ex} \textsf{AttnDin} (orig.)  & 0.213 & 0.318 & 0.131 & 6.515\\
 & \textsf{AttnDin} + CoMOT  & 0.189 & 0.298 & \textbf{0.045}\rlap{$^{\triangledown}$} & 6.666\\
 & \rule{-2pt}{3ex} \textsf{DASALC} (orig.)  & 0.194 & 0.302 & 0.127 & 5.939\\
 & \textsf{DASALC} + CoMOT  & 0.187 & 0.297 & \textbf{0.056}\rlap{$^{\triangledown}$} & 6.115\\

    \bottomrule
    \end{tabular}
\end{table}

\subsection{Comparing against optimal solutions~(\rqtwo{})}
To assess the quality of CoMOT-based ranking policies, we compare them against an optimal solution found using CVX, CVXFOE (optimization problem in Eq.~\ref{eq:linear_program_with_constrains} with constraints E.q.~\ref{eq:demographic_parity_equality}). Therefore, the CVXFOE represents the best fair policy, given explicit constraints. In particular, we use different fairness levels, $ \rho \in [0.01, 0.03, 0.05, 0.1, 0.25, 0.5]$, where $0.5$ and $0.01$ denote small correction and strong fairness correction, respectively (see q.~\ref{eq:linear_program_with_constrains}). 

We do a forward pass on the potential model $f_\theta$ and explicitly compute a fair re-ranking policy for each testing query. Also, we use CVXFOE with a given fairness level $\rho$ to find an optimal fair re-ranking policy. Then, we measure the fairness cost, prediction performance, and computation time for all ranking policies.

\subsubsection{Comparing fairness performance}
\begin{figure*}[!ht]
	%\small
    \footnotesize
	\centering
	{\includegraphics[width=1\linewidth, trim={0 0mm 0 0mm}, clip]{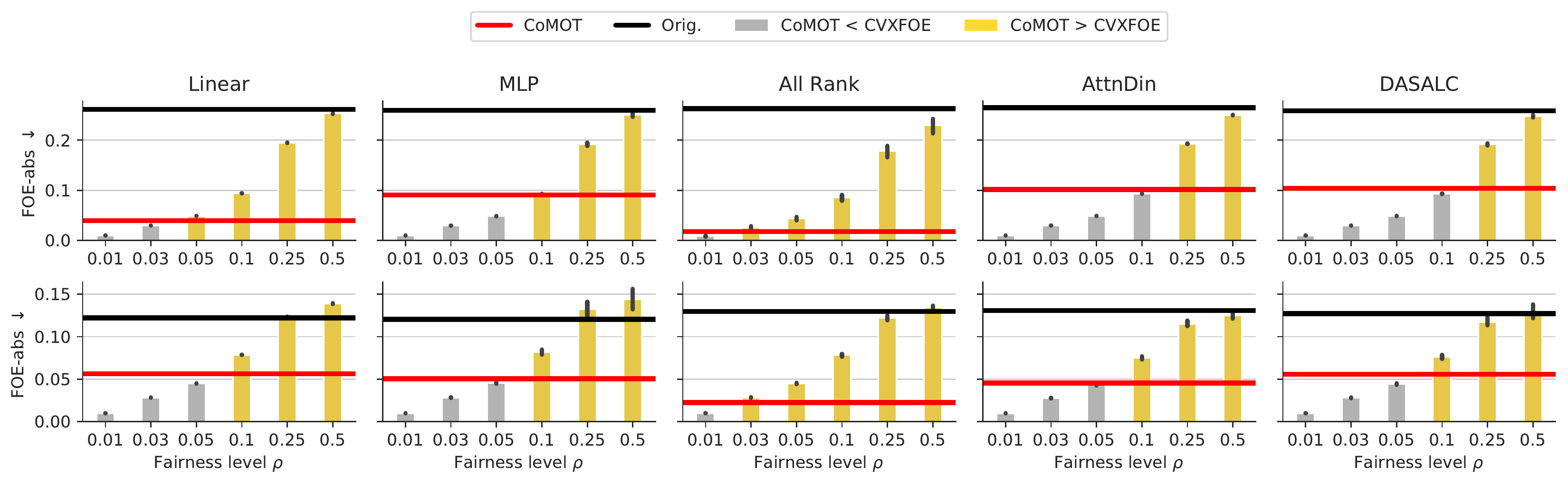}}
	\caption{
Comparing fair ranking policies on TREC 2019 (top row) and TREC 2020 (bottom row) datasets. We compare the fairness performance (the lower, the better) of the original ranking scores (red line) with ranking policies obtained by CoMOT(black line) and by solving a constrained optimization using CVXFOE for various fairness levels $\rho$ (bars). Yellow bars denote the fairness levels where CoMOT outperforms CVXFOE, CoMOT > CVXFOE, and grey bars indicate otherwise.
}
	\label{fig:compare_fairness}
	\Description{XXX}
\end{figure*}

Fig.~\ref{fig:compare_fairness} shows the fairness cost of a policy found using CVXFOE for various fairness levels $\rho$ and different pre-trained rankers. Generally, for most rankers, CoMOT learns to predict a suitable fair-ranking policy equivalent to one found by CVXFOE with a fairness admissibility level of $\rho=0.1$. In particular, on the All Rank ranking scores, CoMOT performs similarly to the stronger fairness constraint tested with CVXFOE, $\rho=0.01$, for the TREC2019 dataset. 

\subsubsection{Comparing computation time}
\begin{figure}[!ht]
	\small
	\centering
	{\includegraphics[width=.8\linewidth, trim={0 0mm 0 0mm}, clip]{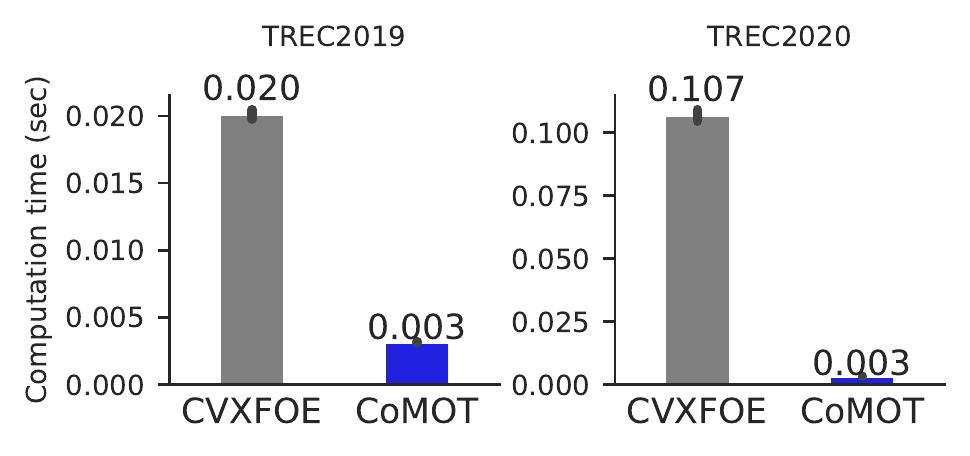}}
	\caption{
Computation time comparison. Getting a ranking policy using CoMOT is 6 and 36 times faster than cold-start optimization using CVXFOE on the TREC 2019 and 2020 datasets, respectively.
 }
	\label{fig:compare_time}
	\Description{Computation time comparison. Getting a ranking policy using CoMOT is 6 and 36 times faster than cold-start optimization using CVX in TREC 2019 and 2020, respectively.}
\end{figure}

CoMOT has quadratic computational complexity during inference due to finetuning using Sinkhorn, which is at least $n$ times less than the complexity for CVX-based solvers. Fig.~\ref{fig:compare_time} shows the computation time on test data for TREC datasets. We observe dataset-dependent gains, approximately 6 and 36 times faster than CVXFOE for TREC 2019 and 2020, respectively. As expected, these gains are in the same order as the average number of documents per query (see Table~\ref{tab:datasets}).

\subsubsection{Comparing ranking performance}
\begin{figure}[!ht]
	\small
	\centering
	{\includegraphics[width=0.95\linewidth, trim={0 0mm 0 0mm}, clip]{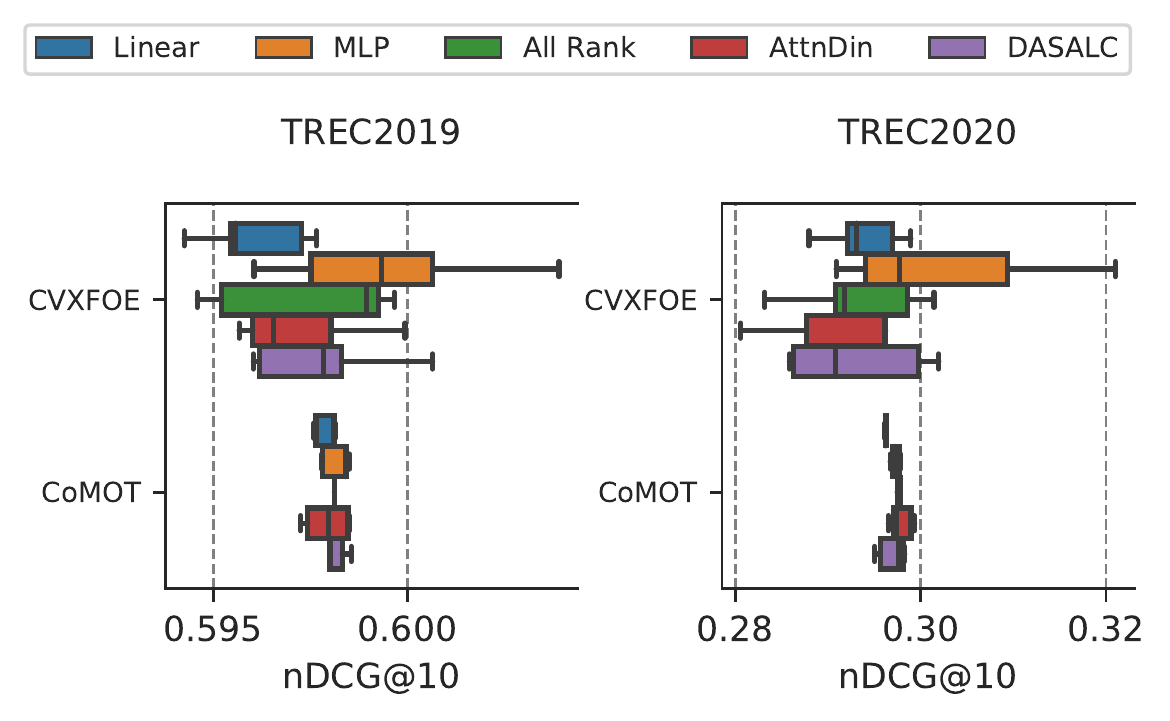}}
	\caption{
 Comparing the ranking performance of re-ranking policies found by CVXFOE and CoMOT for various ranking architectures. We use CVXFOE with a fairness level $\rho = 0.1$. This method is not predictive and denotes the best possible stochastic re-ranking policy. On the other hand, CoMOT is fully predictive and produces similar ranking performance.   
}
	\label{fig:compare_ndcg}
	%\Description{xXX}
\end{figure}

Fig.~\ref{fig:compare_ndcg}. shows the nDCG@10 for ranking policies obtained via CoMOT and CVXFOE with a fairness level $\rho = 0.1$.  
For both datasets, TREC 2019 and 2020,  CoMOT-based policies consistently display lower variance. However, the predicted scores are not statistically different between CVX and CoMOT-based policies.

In summary, CoMOT finds ranking policies comparable to CVX-based solutions in a fraction of the time. This fraction appears to be proportional to the average document list in the dataset. 

\subsection{Sampling from ranking policies~(\rqthree{})}

\subsubsection{Quality of the approximation}
\begin{figure}[!ht]
	\small
	\centering
	{\includegraphics[width=1\linewidth, trim={0 0mm 0 0mm}, clip]{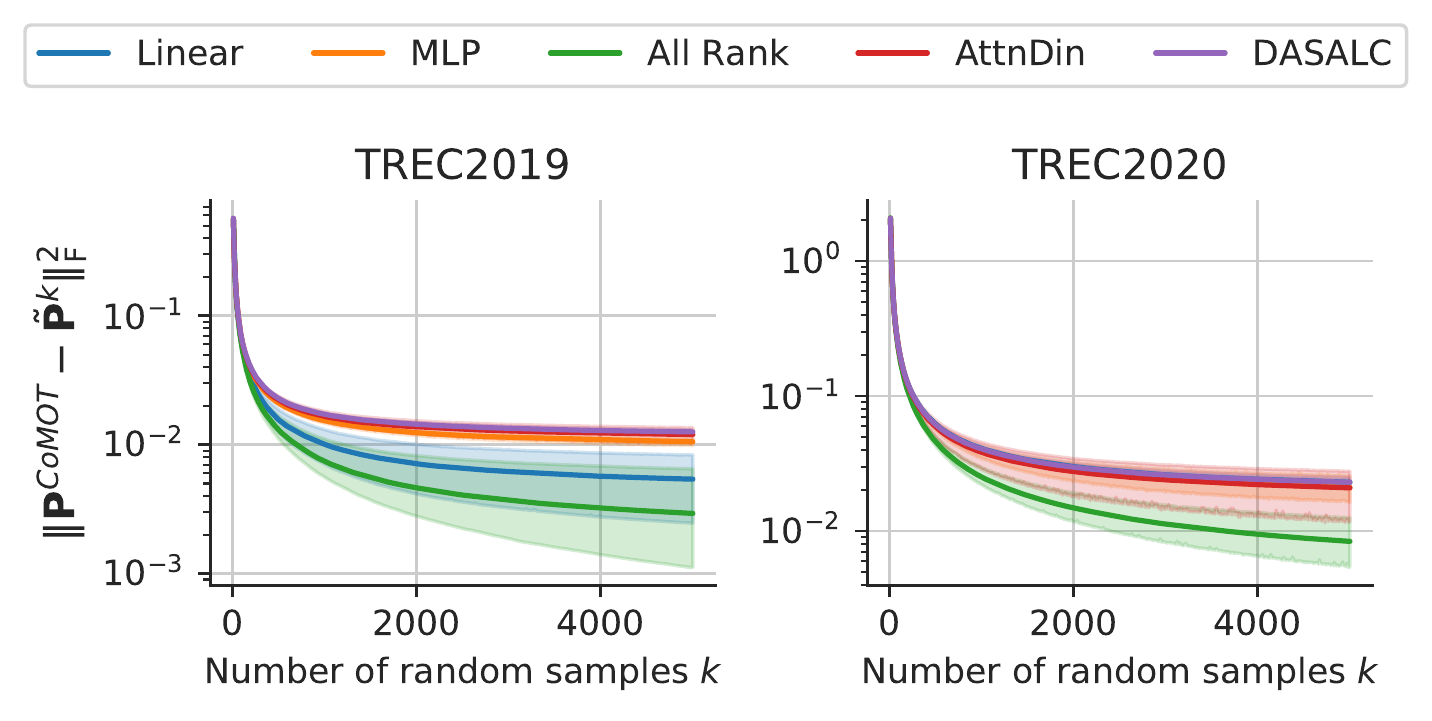}}
	\caption{
Approximating stochastic ranking policies in expectation using GumMS. The average of $k$ GumMS samples, $\tilde{\rmP}^{k}$, approximates a target policy $\rmP^{\CoMOT}$ for a significant enough number of samples $k$ for all rankers.
}
	\label{fig:mc_gumms}
	\Description{XXX}
\end{figure}

Fig.~\ref{fig:mc_gumms} shows the approximation error as a function of the number of drawn samples $k \in [5\,000]$ across five runs. As expected, GumMS approximates the CoMOT policy in expectation for all pre-trained rankers. However, the convergence speed is not the same for all rankers. This speed depends on the GumMS's noise parameter and the policy's sparsity/smoothness. We note that for each query, the number of permutations is $\gO(n_q^2)$ according to the BvND theorem; consequently, the error on the smaller dataset, TREC 2019, displays a lower approximation error for the same number of random draws $k$. 

\subsubsection{Predictive performance}
Table~\ref{tab:sampled} depicts the comparison between the CoMOT ranking policy (ref.) and the average of $k=5\,000$ GumMS samples for such a policy (approx.). For both datasets, sampling using GumMS faithfully represents the original ranking policy in expectation. Generally, the nDCG@10 and optimal transport costs are indistinguishable between the GumMS approximation and the reference. For TREC 2019, the average GumMS samples display significantly lower fairness costs for MLP and DASALC architectures. We observe similar behavior in the TREC 2020 dataset for the Linear ranker. However, we hypothesize these effects are artifacts and will disappear for $k \rightarrow \infty$.
\begin{table}[!ht]
    \small
    \centering
    \caption{
Quality of the GumMS's policy approximation for various ranking architectures. We use a two-tailed paired t-test ($p < 0.01$) to compare each CoMOT policy \emph{ref.} with its empirical expectation via GumMS ($k=5\,000$). Other conventions are the same as in Table~\ref{tab:prediction}.
}
    \label{tab:sampled}
    \begin{tabular}{cl@{\hspace{3mm}}rrrrr}\\
    \toprule
    & \multirow{2}{*}{Ranking policy} & \multicolumn{2}{c}{\textsf{nDCG$\uparrow$}} & Fairness$\downarrow$  & Utility $\uparrow$ \\
    \cmidrule(lr){3-4} \cmidrule(lr){5-5} \cmidrule(lr){6-6}
     & & @5 & @10 & FOE abs &  Neg. OT cost  \\
    \midrule
    \parbox[t]{2mm}{\multirow{10}{*}{\rotatebox[origin=c]{90}{\textsf{TREC2019}}}}  & \textsf{Linear} + CoMOT (ref.) & 0.512 & 0.598 & 0.039 & 1.991\\
 & \textsf{Linear} + GumMS (approx.)  & 0.512 & 0.598 & 0.031 & 1.980\\
 & \rule{-2pt}{3ex} \textsf{MLP} + CoMOT (ref.)  & 0.512 & 0.598 & 0.091 & 2.722\\
 & \textsf{MLP} + GumMS (approx.)  & 0.512 & 0.598 & 0.070\rlap{$^{\triangledown}$} & 2.707\\
 & \rule{-2pt}{3ex} \textsf{All Rank} + CoMOT (ref.)  & 0.512 & 0.598 & 0.018 & 2.506\\
 & \textsf{All Rank} + GumMS (approx.)  & 0.512 & 0.598 & 0.016 & 2.502\\
 & \rule{-2pt}{3ex} \textsf{AttnDin} + CoMOT (ref)  & 0.512 & 0.598 & 0.101 & 2.935\\
 & \textsf{AttnDin} + GumMS (approx.)  & 0.512 & 0.598 & 0.079 & 2.921\\
 & \rule{-2pt}{3ex} \textsf{DASALC} + CoMOT (ref.)  & 0.512 & 0.598 & 0.104 & 2.952\\
 & \textsf{DASALC} + GumMS (approx.)  & 0.512 & 0.598 & 0.081\rlap{$^{\triangledown}$} & 2.937\\
\midrule
\parbox[t]{2mm}{\multirow{10}{*}{\rotatebox[origin=c]{90}{\textsf{TREC2020}}}}  & \textsf{Linear} + CoMOT (ref.) & \textbf{0.188} & 0.296 & 0.056 & 6.386\\
 & \textsf{Linear} + GumMS (approx.)  & 0.188\rlap{$^{\triangledown}$} & 0.296 & 0.054\rlap{$^{\triangledown}$} & 6.391\\
 & \rule{-2pt}{3ex} \textsf{MLP} + CoMOT (ref.)  & 0.188 & 0.297 & 0.050 & 5.525\\
 & \textsf{MLP} + GumMS (approx.)  & 0.188 & 0.297 & 0.047 & 5.527\\
 & \rule{-2pt}{3ex} \textsf{All Rank} + CoMOT (ref.)  & 0.188 & 0.298 & 0.022 & 5.399\\
 & \textsf{All Rank} + GumMS (approx.)  & 0.188 & 0.298 & 0.018 & 5.375\\
 & \rule{-2pt}{3ex} \textsf{AttnDin} + CoMOT (ref.)  & 0.189 & 0.298 & 0.045 & 6.666\\
 & \textsf{AttnDin} + GumMS (approx.)  & 0.189 & 0.298 & 0.045 & 6.672\\
 & \rule{-2pt}{3ex} \textsf{DASALC} + CoMOT (ref.)  & 0.187 & 0.297 & 0.056 & 6.115\\
 & \textsf{DASALC} + GumMS (approx.)  & 0.187 & 0.297 & 0.053 & 6.118\\

    \bottomrule
    \end{tabular}
\end{table}

\section{Conclusion}
We introduced Constrained Meta-Optimal Transport (CoMOT), an approach designed to predict stochastic fair re-ranking policies based on Meta-Optimal Transport. CoMOT leverages the repetitive nature of acquiring re-ranking policies for each query by learning to predict approximately optimal constrained policies. As a post-processing method, CoMOT is independent of the ranking architecture and loss. However, predicting the policy is insufficient, as we need to sample rankings from such a policy during deployment. To address this, we introduced Gumbel-Matching Sampling (GumMS), an online method to draw rankings at random from a stochastic ranking policy encoded as a Doubly-Stochastic (DS) matrix.

We empirically verified that CoMOT leverages the training data as it generalizes to unseen queries. The CoMOT-based re-ranking policies always displayed a lower fairness cost than the original policy. Also, it had a similar ranking and fairness performance as the solution to the original constrained optimization problem at a fraction of the computation time, between 6 and 36 times faster in TREC 2019 and 2020, respectively. 

Our experiments showed GumMS's usefulness for sampling rankings from DS-based policies. However, GumMS has a quadratic computation complexity in the number of documents, similar to transformer-based architectures~\cite{vaswani2017attention}. Thus, it needs to control the number of documents by pre-selecting them to avoid computation bottlenecks.
    
\paragraph{Limitations and future work}
Our proposed pipeline is well-suited in cases where the number of queries is much larger than the number of documents per query. Primarily, CoMOT with GumMS handles new queries in an online fashion. 
This work focuses on re-ranking using relevance-like scores. Alternatively,  CoMOT could also use document features to improve generalization. Also, we need to theoretically define the effect of the noise value on GumMS concentration performance. Finally,  we compared CoMOT to a highly optimized but general-purpose solver, CVX. We could rely on custom-built solvers. However, we wanted to showcase the versatility of CoMOT as it can easily include non-convex but (sub)differentiable constraints as a learning objective.

%\section{Acknowledgments}

\bibliographystyle{ACM-Reference-Format}
\bibliography{biblio}


\begin{thebibliography}{61}

%%% the Web address.

\ifx \showCODEN    \undefined \def \showCODEN     #1{\unskip}     \fi
\ifx \showDOI      \undefined \def \showDOI       #1{#1}\fi
\ifx \showISBNx    \undefined \def \showISBNx     #1{\unskip}     \fi
\ifx \showISBNxiii \undefined \def \showISBNxiii  #1{\unskip}     \fi
\ifx \showISSN     \undefined \def \showISSN      #1{\unskip}     \fi
\ifx \showLCCN     \undefined \def \showLCCN      #1{\unskip}     \fi
\ifx \shownote     \undefined \def \shownote      #1{#1}          \fi
\ifx \showarticletitle \undefined \def \showarticletitle #1{#1}   \fi
\ifx \showURL      \undefined \def \showURL       {\relax}        \fi
% invisible to TeX
\providecommand\bibfield[2]{#2}
\providecommand\bibinfo[2]{#2}
\providecommand\natexlab[1]{#1}
\providecommand\showeprint[2][]{arXiv:#2}

\bibitem[Abdollahpouri(2019)]%
        {abdollahpouri2019popularity}
\bibfield{author}{\bibinfo{person}{Himan Abdollahpouri}.}
  \bibinfo{year}{2019}\natexlab{}.
\newblock \showarticletitle{Popularity bias in ranking and recommendation}. In
  \bibinfo{booktitle}{\emph{AAAI}}. \bibinfo{pages}{529--530}.
\newblock


\bibitem[Adams and Zemel(2011)]%
        {adams2011ranking}
\bibfield{author}{\bibinfo{person}{Ryan~Prescott Adams} {and}
  \bibinfo{person}{Richard~S Zemel}.} \bibinfo{year}{2011}\natexlab{}.
\newblock \showarticletitle{Ranking via sinkhorn propagation}.
\newblock \bibinfo{journal}{\emph{arXiv preprint arXiv:1106.1925}}
  (\bibinfo{year}{2011}).
\newblock


\bibitem[Agarwal et~al\mbox{.}(2019)]%
        {agarwal2019general}
\bibfield{author}{\bibinfo{person}{Aman Agarwal}, \bibinfo{person}{Kenta
  Takatsu}, \bibinfo{person}{Ivan Zaitsev}, {and} \bibinfo{person}{Thorsten
  Joachims}.} \bibinfo{year}{2019}\natexlab{}.
\newblock \showarticletitle{A general framework for counterfactual
  learning-to-rank}. In \bibinfo{booktitle}{\emph{ACM SIGIR}}.
  \bibinfo{pages}{5--14}.
\newblock


\bibitem[Amos(2022)]%
        {amos2022tutorial}
\bibfield{author}{\bibinfo{person}{Brandon Amos}.}
  \bibinfo{year}{2022}\natexlab{}.
\newblock \showarticletitle{Tutorial on amortized optimization for learning to
  optimize over continuous domains}.
\newblock \bibinfo{journal}{\emph{arXiv preprint arXiv:2202.00665}}
  (\bibinfo{year}{2022}).
\newblock


\bibitem[Amos et~al\mbox{.}(2022)]%
        {amos2022meta}
\bibfield{author}{\bibinfo{person}{Brandon Amos}, \bibinfo{person}{Samuel
  Cohen}, \bibinfo{person}{Giulia Luise}, {and} \bibinfo{person}{Ievgen
  Redko}.} \bibinfo{year}{2022}\natexlab{}.
\newblock \showarticletitle{Meta optimal transport}.
\newblock \bibinfo{journal}{\emph{arXiv preprint arXiv:2206.05262}}
  (\bibinfo{year}{2022}).
\newblock


\bibitem[Asudeh et~al\mbox{.}(2019)]%
        {asudeh2019designing}
\bibfield{author}{\bibinfo{person}{Abolfazl Asudeh}, \bibinfo{person}{HV
  Jagadish}, \bibinfo{person}{Julia Stoyanovich}, {and} \bibinfo{person}{Gautam
  Das}.} \bibinfo{year}{2019}\natexlab{}.
\newblock \showarticletitle{Designing fair ranking schemes}. In
  \bibinfo{booktitle}{\emph{SIGMOD}}. \bibinfo{pages}{1259--1276}.
\newblock


\bibitem[Biega et~al\mbox{.}(2018)]%
        {biega2018equity}
\bibfield{author}{\bibinfo{person}{Asia~J Biega}, \bibinfo{person}{Krishna~P
  Gummadi}, {and} \bibinfo{person}{Gerhard Weikum}.}
  \bibinfo{year}{2018}\natexlab{}.
\newblock \showarticletitle{Equity of attention: Amortizing individual fairness
  in rankings}. In \bibinfo{booktitle}{\emph{ACM SIGIR}}.
  \bibinfo{pages}{405--414}.
\newblock


\bibitem[Birkhoff(1940)]%
        {birkhoff1940lattice}
\bibfield{author}{\bibinfo{person}{Garrett Birkhoff}.}
  \bibinfo{year}{1940}\natexlab{}.
\newblock \bibinfo{booktitle}{\emph{Lattice theory}}.
  Vol.~\bibinfo{volume}{25}.
\newblock \bibinfo{publisher}{American Mathematical Soc.}
\newblock


\bibitem[Birkhoff(1946)]%
        {birkhoff1946tres}
\bibfield{author}{\bibinfo{person}{Garrett Birkhoff}.}
  \bibinfo{year}{1946}\natexlab{}.
\newblock \showarticletitle{Tres observaciones sobre el algebra lineal}.
\newblock \bibinfo{journal}{\emph{Univ. Nac. Tucuman, Ser. A}}
  \bibinfo{volume}{5} (\bibinfo{year}{1946}), \bibinfo{pages}{147--154}.
\newblock


\bibitem[Bruch et~al\mbox{.}(2020)]%
        {bruch2020stochastic}
\bibfield{author}{\bibinfo{person}{Sebastian Bruch}, \bibinfo{person}{Shuguang
  Han}, \bibinfo{person}{Michael Bendersky}, {and} \bibinfo{person}{Marc
  Najork}.} \bibinfo{year}{2020}\natexlab{}.
\newblock \showarticletitle{A stochastic treatment of learning to rank scoring
  functions}. In \bibinfo{booktitle}{\emph{ACM WSDM}}. \bibinfo{pages}{61--69}.
\newblock


\bibitem[Cao et~al\mbox{.}(2007)]%
        {cao2007learning}
\bibfield{author}{\bibinfo{person}{Zhe Cao}, \bibinfo{person}{Tao Qin},
  \bibinfo{person}{Tie-Yan Liu}, \bibinfo{person}{Ming-Feng Tsai}, {and}
  \bibinfo{person}{Hang Li}.} \bibinfo{year}{2007}\natexlab{}.
\newblock \showarticletitle{Learning to rank: from pairwise approach to
  listwise approach}. In \bibinfo{booktitle}{\emph{ICML}}.
  \bibinfo{pages}{129--136}.
\newblock


\bibitem[Celis et~al\mbox{.}(2017)]%
        {celis2017ranking}
\bibfield{author}{\bibinfo{person}{L~Elisa Celis}, \bibinfo{person}{Damian
  Straszak}, {and} \bibinfo{person}{Nisheeth~K Vishnoi}.}
  \bibinfo{year}{2017}\natexlab{}.
\newblock \showarticletitle{Ranking with fairness constraints}.
\newblock \bibinfo{journal}{\emph{arXiv preprint arXiv:1704.06840}}
  (\bibinfo{year}{2017}).
\newblock


\bibitem[Chapelle and Zhang(2009)]%
        {chapelle2009dynamic}
\bibfield{author}{\bibinfo{person}{Olivier Chapelle} {and} \bibinfo{person}{Ya
  Zhang}.} \bibinfo{year}{2009}\natexlab{}.
\newblock \showarticletitle{A dynamic bayesian network click model for web
  search ranking}. In \bibinfo{booktitle}{\emph{WWW}}. \bibinfo{pages}{1--10}.
\newblock


\bibitem[Crawford(2017)]%
        {crawford2017trouble}
\bibfield{author}{\bibinfo{person}{Kate Crawford}.}
  \bibinfo{year}{2017}\natexlab{}.
\newblock \showarticletitle{The trouble with bias}. In
  \bibinfo{booktitle}{\emph{NeurIPS, invited speaker}}.
\newblock


\bibitem[Cuturi(2013)]%
        {cuturi2013sinkhorn}
\bibfield{author}{\bibinfo{person}{Marco Cuturi}.}
  \bibinfo{year}{2013}\natexlab{}.
\newblock \showarticletitle{Sinkhorn distances: Lightspeed computation of
  optimal transport}.
\newblock \bibinfo{journal}{\emph{NeurIPS}}  \bibinfo{volume}{26}
  (\bibinfo{year}{2013}).
\newblock


\bibitem[Cuturi et~al\mbox{.}(2019)]%
        {cuturi2019differentiable}
\bibfield{author}{\bibinfo{person}{Marco Cuturi}, \bibinfo{person}{Olivier
  Teboul}, {and} \bibinfo{person}{Jean-Philippe Vert}.}
  \bibinfo{year}{2019}\natexlab{}.
\newblock \showarticletitle{Differentiable ranking and sorting using optimal
  transport}.
\newblock \bibinfo{journal}{\emph{NeurIPS}}  \bibinfo{volume}{32}
  (\bibinfo{year}{2019}).
\newblock


\bibitem[Diamond and Boyd(2016)]%
        {diamond2016cvxpy}
\bibfield{author}{\bibinfo{person}{Steven Diamond} {and}
  \bibinfo{person}{Stephen Boyd}.} \bibinfo{year}{2016}\natexlab{}.
\newblock \showarticletitle{{CVXPY}: {A} {P}ython-embedded modeling language
  for convex optimization}.
\newblock \bibinfo{journal}{\emph{JMLR}} \bibinfo{volume}{17},
  \bibinfo{number}{83} (\bibinfo{year}{2016}), \bibinfo{pages}{1--5}.
\newblock


\bibitem[Diaz et~al\mbox{.}(2020)]%
        {diaz2020evaluating}
\bibfield{author}{\bibinfo{person}{Fernando Diaz}, \bibinfo{person}{Bhaskar
  Mitra}, \bibinfo{person}{Michael~D Ekstrand}, \bibinfo{person}{Asia~J Biega},
  {and} \bibinfo{person}{Ben Carterette}.} \bibinfo{year}{2020}\natexlab{}.
\newblock \showarticletitle{Evaluating stochastic rankings with expected
  exposure}. In \bibinfo{booktitle}{\emph{ACM CIKM}}.
  \bibinfo{pages}{275--284}.
\newblock


\bibitem[Do and Usunier(2022)]%
        {do2022optimizing}
\bibfield{author}{\bibinfo{person}{Virginie Do} {and} \bibinfo{person}{Nicolas
  Usunier}.} \bibinfo{year}{2022}\natexlab{}.
\newblock \showarticletitle{Optimizing generalized Gini indices for fairness in
  rankings}.
\newblock \bibinfo{journal}{\emph{ACM SIGIR}} (\bibinfo{year}{2022}),
  \bibinfo{pages}{737--747}.
\newblock


\bibitem[Dufoss{\'e} et~al\mbox{.}(2018)]%
        {dufosse2018further}
\bibfield{author}{\bibinfo{person}{Fanny Dufoss{\'e}}, \bibinfo{person}{Kamer
  Kaya}, \bibinfo{person}{Ioannis Panagiotas}, {and} \bibinfo{person}{Bora
  U{\c{c}}ar}.} \bibinfo{year}{2018}\natexlab{}.
\newblock \showarticletitle{Further notes on Birkhoff--von Neumann
  decomposition of doubly stochastic matrices}.
\newblock \bibinfo{journal}{\emph{Linear Algebra Appl.}}  \bibinfo{volume}{554}
  (\bibinfo{year}{2018}), \bibinfo{pages}{68--78}.
\newblock


\bibitem[Dufoss{\'e} and U{\c{c}}ar(2016)]%
        {dufosse2016notes}
\bibfield{author}{\bibinfo{person}{Fanny Dufoss{\'e}} {and}
  \bibinfo{person}{Bora U{\c{c}}ar}.} \bibinfo{year}{2016}\natexlab{}.
\newblock \showarticletitle{Notes on Birkhoff--von Neumann decomposition of
  doubly stochastic matrices}.
\newblock \bibinfo{journal}{\emph{Linear Algebra Appl.}}  \bibinfo{volume}{497}
  (\bibinfo{year}{2016}), \bibinfo{pages}{108--115}.
\newblock


\bibitem[Ekstrand et~al\mbox{.}(2022)]%
        {ekstrand2022fairness}
\bibfield{author}{\bibinfo{person}{Michael~D Ekstrand},
  \bibinfo{person}{Anubrata Das}, \bibinfo{person}{Robin Burke},
  \bibinfo{person}{Fernando Diaz}, {et~al\mbox{.}}}
  \bibinfo{year}{2022}\natexlab{}.
\newblock \showarticletitle{Fairness in information access systems}.
\newblock \bibinfo{journal}{\emph{Foundations and Trends{\textregistered} in
  Information Retrieval}} \bibinfo{volume}{16}, \bibinfo{number}{1-2}
  (\bibinfo{year}{2022}), \bibinfo{pages}{1--177}.
\newblock


\bibitem[Flamary et~al\mbox{.}(2021)]%
        {flamary2021pot}
\bibfield{author}{\bibinfo{person}{R{\'e}mi Flamary}, \bibinfo{person}{Nicolas
  Courty}, \bibinfo{person}{Alexandre Gramfort}, \bibinfo{person}{Mokhtar~Z.
  Alaya}, \bibinfo{person}{Aur{\'e}lie Boisbunon}, \bibinfo{person}{Stanislas
  Chambon}, \bibinfo{person}{Laetitia Chapel}, \bibinfo{person}{Adrien
  Corenflos}, \bibinfo{person}{Kilian Fatras}, \bibinfo{person}{Nemo Fournier},
  \bibinfo{person}{L{\'e}o Gautheron}, \bibinfo{person}{Nathalie~T.H. Gayraud},
  \bibinfo{person}{Hicham Janati}, \bibinfo{person}{Alain Rakotomamonjy},
  \bibinfo{person}{Ievgen Redko}, \bibinfo{person}{Antoine Rolet},
  \bibinfo{person}{Antony Schutz}, \bibinfo{person}{Vivien Seguy},
  \bibinfo{person}{Danica~J. Sutherland}, \bibinfo{person}{Romain Tavenard},
  \bibinfo{person}{Alexander Tong}, {and} \bibinfo{person}{Titouan Vayer}.}
  \bibinfo{year}{2021}\natexlab{}.
\newblock \showarticletitle{POT: Python Optimal Transport}.
\newblock \bibinfo{journal}{\emph{JMLR}} \bibinfo{volume}{22},
  \bibinfo{number}{78} (\bibinfo{year}{2021}), \bibinfo{pages}{1--8}.
\newblock
\urldef\tempurl%
\url{http://jmlr.org/papers/v22/20-451.html}
\showURL{%
\tempurl}


\bibitem[Friedman and Nissenbaum(1996)]%
        {friedman1996bias}
\bibfield{author}{\bibinfo{person}{Batya Friedman} {and} \bibinfo{person}{Helen
  Nissenbaum}.} \bibinfo{year}{1996}\natexlab{}.
\newblock \showarticletitle{Bias in computer systems}.
\newblock \bibinfo{journal}{\emph{ACM TOIS}} \bibinfo{volume}{14},
  \bibinfo{number}{3} (\bibinfo{year}{1996}), \bibinfo{pages}{330--347}.
\newblock


\bibitem[Gane et~al\mbox{.}(2014)]%
        {gane2014learning}
\bibfield{author}{\bibinfo{person}{Andreea Gane}, \bibinfo{person}{Tamir
  Hazan}, {and} \bibinfo{person}{Tommi Jaakkola}.}
  \bibinfo{year}{2014}\natexlab{}.
\newblock \showarticletitle{Learning with maximum a-posteriori perturbation
  models}. In \bibinfo{booktitle}{\emph{AISTATS}}. PMLR,
  \bibinfo{pages}{247--256}.
\newblock


\bibitem[Gao and Shah(2020)]%
        {gao2020toward}
\bibfield{author}{\bibinfo{person}{Ruoyuan Gao} {and} \bibinfo{person}{Chirag
  Shah}.} \bibinfo{year}{2020}\natexlab{}.
\newblock \showarticletitle{Toward creating a fairer ranking in search engine
  results}.
\newblock \bibinfo{journal}{\emph{Information Processing \& Management}}
  \bibinfo{volume}{57}, \bibinfo{number}{1} (\bibinfo{year}{2020}),
  \bibinfo{pages}{102138}.
\newblock


\bibitem[Heuss et~al\mbox{.}(2022)]%
        {heuss2022fairness}
\bibfield{author}{\bibinfo{person}{Maria Heuss}, \bibinfo{person}{Fatemeh
  Sarvi}, {and} \bibinfo{person}{Maarten de Rijke}.}
  \bibinfo{year}{2022}\natexlab{}.
\newblock \showarticletitle{Fairness of Exposure in Light of Incomplete
  Exposure Estimation}.
\newblock \bibinfo{journal}{\emph{ACM SIGIR}} (\bibinfo{year}{2022}),
  \bibinfo{pages}{759--769}.
\newblock


\bibitem[Jagerman and de~Rijke(2020)]%
        {jagerman2020accelerated}
\bibfield{author}{\bibinfo{person}{Rolf Jagerman} {and}
  \bibinfo{person}{Maarten de Rijke}.} \bibinfo{year}{2020}\natexlab{}.
\newblock \showarticletitle{Accelerated Convergence for Counterfactual Learning
  to Rank}. In \bibinfo{booktitle}{\emph{ACM SIGIR}}.
  \bibinfo{pages}{469--478}.
\newblock


\bibitem[Jang et~al\mbox{.}(2017)]%
        {jang2016categorical}
\bibfield{author}{\bibinfo{person}{Eric Jang}, \bibinfo{person}{Shixiang Gu},
  {and} \bibinfo{person}{Ben Poole}.} \bibinfo{year}{2017}\natexlab{}.
\newblock \showarticletitle{Categorical reparameterization with
  gumbel-softmax}.
\newblock \bibinfo{journal}{\emph{ICLR}} (\bibinfo{year}{2017}).
\newblock


\bibitem[Joachims(2002)]%
        {joachims2002optimizing}
\bibfield{author}{\bibinfo{person}{Thorsten Joachims}.}
  \bibinfo{year}{2002}\natexlab{}.
\newblock \showarticletitle{Optimizing search engines using clickthrough data}.
  In \bibinfo{booktitle}{\emph{ACM SIGIR}}. \bibinfo{pages}{133--142}.
\newblock


\bibitem[Kay et~al\mbox{.}(2015)]%
        {kay2015unequal}
\bibfield{author}{\bibinfo{person}{Matthew Kay}, \bibinfo{person}{Cynthia
  Matuszek}, {and} \bibinfo{person}{Sean~A Munson}.}
  \bibinfo{year}{2015}\natexlab{}.
\newblock \showarticletitle{Unequal representation and gender stereotypes in
  image search results for occupations}. In \bibinfo{booktitle}{\emph{CHI}}.
  \bibinfo{pages}{3819--3828}.
\newblock


\bibitem[Kletti et~al\mbox{.}(2022)]%
        {kletti2022introducing}
\bibfield{author}{\bibinfo{person}{Till Kletti}, \bibinfo{person}{Jean-Michel
  Renders}, {and} \bibinfo{person}{Patrick Loiseau}.}
  \bibinfo{year}{2022}\natexlab{}.
\newblock \showarticletitle{Introducing the Expohedron for Efficient
  Pareto-optimal Fairness-Utility Amortizations in Repeated Rankings}. In
  \bibinfo{booktitle}{\emph{ACM WSDM}}. \bibinfo{pages}{498--507}.
\newblock


\bibitem[Kotary et~al\mbox{.}(2022)]%
        {kotary2022end}
\bibfield{author}{\bibinfo{person}{James Kotary}, \bibinfo{person}{Ferdinando
  Fioretto}, \bibinfo{person}{Pascal Van~Hentenryck}, {and}
  \bibinfo{person}{Ziwei Zhu}.} \bibinfo{year}{2022}\natexlab{}.
\newblock \showarticletitle{End-to-End Learning for Fair Ranking Systems}. In
  \bibinfo{booktitle}{\emph{WWW}}. \bibinfo{pages}{3520--3530}.
\newblock


\bibitem[Kuhn(1955)]%
        {kuhn1955hungarian}
\bibfield{author}{\bibinfo{person}{Harold~W Kuhn}.}
  \bibinfo{year}{1955}\natexlab{}.
\newblock \showarticletitle{The Hungarian method for the assignment problem}.
\newblock \bibinfo{journal}{\emph{Naval research logistics quarterly}}
  \bibinfo{volume}{2}, \bibinfo{number}{1-2} (\bibinfo{year}{1955}),
  \bibinfo{pages}{83--97}.
\newblock


\bibitem[Loshchilov and Hutter(2018)]%
        {loshchilov2018decoupled}
\bibfield{author}{\bibinfo{person}{Ilya Loshchilov} {and}
  \bibinfo{person}{Frank Hutter}.} \bibinfo{year}{2018}\natexlab{}.
\newblock \showarticletitle{Decoupled Weight Decay Regularization}. In
  \bibinfo{booktitle}{\emph{ICLR}}.
\newblock


\bibitem[Maddison et~al\mbox{.}(2017)]%
        {maddison2016concrete}
\bibfield{author}{\bibinfo{person}{Chris~J Maddison}, \bibinfo{person}{Andriy
  Mnih}, {and} \bibinfo{person}{Yee~Whye Teh}.}
  \bibinfo{year}{2017}\natexlab{}.
\newblock \showarticletitle{The concrete distribution: A continuous relaxation
  of discrete random variables}.
\newblock \bibinfo{journal}{\emph{ICLR}} (\bibinfo{year}{2017}).
\newblock


\bibitem[Mena et~al\mbox{.}(2018)]%
        {mena2018learning}
\bibfield{author}{\bibinfo{person}{Gonzalo Mena}, \bibinfo{person}{David
  Belanger}, \bibinfo{person}{Scott Linderman}, {and} \bibinfo{person}{Jasper
  Snoek}.} \bibinfo{year}{2018}\natexlab{}.
\newblock \showarticletitle{Learning latent permutations with gumbel-sinkhorn
  networks}.
\newblock \bibinfo{journal}{\emph{ICLR}} (\bibinfo{year}{2018}).
\newblock


\bibitem[Oosterhuis(2021)]%
        {oosterhuis2021computationally}
\bibfield{author}{\bibinfo{person}{Harrie Oosterhuis}.}
  \bibinfo{year}{2021}\natexlab{}.
\newblock \showarticletitle{Computationally efficient optimization of
  plackett-luce ranking models for relevance and fairness}. In
  \bibinfo{booktitle}{\emph{ACM SIGIR}}. \bibinfo{pages}{1023--1032}.
\newblock


\bibitem[Pandey et~al\mbox{.}(2005)]%
        {pandey2005shuffling}
\bibfield{author}{\bibinfo{person}{Sandeep Pandey}, \bibinfo{person}{Sourashis
  Roy}, \bibinfo{person}{Christopher Olston}, \bibinfo{person}{Junghoo Cho},
  {and} \bibinfo{person}{Soumen Chakrabarti}.} \bibinfo{year}{2005}\natexlab{}.
\newblock \showarticletitle{Shuffling a Stacked Deck: The Case for Partially
  Randomized Ranking of Search Engine Results}. In
  \bibinfo{booktitle}{\emph{VLDB}}. \bibinfo{pages}{781–792}.
\newblock


\bibitem[Pasumarthi et~al\mbox{.}(2020)]%
        {pasumarthi2020permutation}
\bibfield{author}{\bibinfo{person}{Rama~Kumar Pasumarthi},
  \bibinfo{person}{Honglei Zhuang}, \bibinfo{person}{Xuanhui Wang},
  \bibinfo{person}{Michael Bendersky}, {and} \bibinfo{person}{Marc Najork}.}
  \bibinfo{year}{2020}\natexlab{}.
\newblock \showarticletitle{Permutation equivariant document interaction
  network for neural learning to rank}. In \bibinfo{booktitle}{\emph{ACM
  SIGIR}}. \bibinfo{pages}{145--148}.
\newblock


\bibitem[Paszke et~al\mbox{.}(2019)]%
        {NEURIPS2019_9015}
\bibfield{author}{\bibinfo{person}{Adam Paszke}, \bibinfo{person}{Sam Gross},
  \bibinfo{person}{Francisco Massa}, \bibinfo{person}{Adam Lerer},
  \bibinfo{person}{James Bradbury}, \bibinfo{person}{Gregory Chanan},
  \bibinfo{person}{Trevor Killeen}, \bibinfo{person}{Zeming Lin},
  \bibinfo{person}{Natalia Gimelshein}, \bibinfo{person}{Luca Antiga},
  \bibinfo{person}{Alban Desmaison}, \bibinfo{person}{Andreas Kopf},
  \bibinfo{person}{Edward Yang}, \bibinfo{person}{Zachary DeVito},
  \bibinfo{person}{Martin Raison}, \bibinfo{person}{Alykhan Tejani},
  \bibinfo{person}{Sasank Chilamkurthy}, \bibinfo{person}{Benoit Steiner},
  \bibinfo{person}{Lu Fang}, \bibinfo{person}{Junjie Bai}, {and}
  \bibinfo{person}{Soumith Chintala}.} \bibinfo{year}{2019}\natexlab{}.
\newblock \showarticletitle{PyTorch: An Imperative Style, High-Performance Deep
  Learning Library}.
\newblock In \bibinfo{booktitle}{\emph{NeurIPS}}. \bibinfo{pages}{8024--8035}.
\newblock


\bibitem[Peyr{\'e} et~al\mbox{.}(2019)]%
        {peyre2019computational}
\bibfield{author}{\bibinfo{person}{Gabriel Peyr{\'e}}, \bibinfo{person}{Marco
  Cuturi}, {et~al\mbox{.}}} \bibinfo{year}{2019}\natexlab{}.
\newblock \showarticletitle{Computational optimal transport: With applications
  to data science}.
\newblock \bibinfo{journal}{\emph{Foundations and Trends{\textregistered} in
  Machine Learning}} \bibinfo{volume}{11}, \bibinfo{number}{5-6}
  (\bibinfo{year}{2019}), \bibinfo{pages}{355--607}.
\newblock


\bibitem[Pobrotyn et~al\mbox{.}(2020)]%
        {pobrotyn2020context}
\bibfield{author}{\bibinfo{person}{Przemys{\l}aw Pobrotyn},
  \bibinfo{person}{Tomasz Bartczak}, \bibinfo{person}{Miko{\l}aj Synowiec},
  \bibinfo{person}{Rados{\l}aw Bia{\l}obrzeski}, {and}
  \bibinfo{person}{Jaros{\l}aw Bojar}.} \bibinfo{year}{2020}\natexlab{}.
\newblock \showarticletitle{Context-aware learning to rank with
  self-attention}.
\newblock \bibinfo{journal}{\emph{arXiv preprint arXiv:2005.10084}}
  (\bibinfo{year}{2020}).
\newblock


\bibitem[Qin et~al\mbox{.}(2020)]%
        {qin2020neural}
\bibfield{author}{\bibinfo{person}{Zhen Qin}, \bibinfo{person}{Le Yan},
  \bibinfo{person}{Honglei Zhuang}, \bibinfo{person}{Yi Tay},
  \bibinfo{person}{Rama~Kumar Pasumarthi}, \bibinfo{person}{Xuanhui Wang},
  \bibinfo{person}{Michael Bendersky}, {and} \bibinfo{person}{Marc Najork}.}
  \bibinfo{year}{2020}\natexlab{}.
\newblock \showarticletitle{Are Neural Rankers still Outperformed by Gradient
  Boosted Decision Trees?}. In \bibinfo{booktitle}{\emph{ICLR}}.
\newblock


\bibitem[Radlinski et~al\mbox{.}(2008)]%
        {radlinski2008learning}
\bibfield{author}{\bibinfo{person}{Filip Radlinski}, \bibinfo{person}{Robert
  Kleinberg}, {and} \bibinfo{person}{Thorsten Joachims}.}
  \bibinfo{year}{2008}\natexlab{}.
\newblock \showarticletitle{Learning diverse rankings with multi-armed
  bandits}. In \bibinfo{booktitle}{\emph{ICML}}. \bibinfo{pages}{784--791}.
\newblock


\bibitem[Robertson et~al\mbox{.}(2009)]%
        {robertson2009probabilistic}
\bibfield{author}{\bibinfo{person}{Stephen Robertson}, \bibinfo{person}{Hugo
  Zaragoza}, {et~al\mbox{.}}} \bibinfo{year}{2009}\natexlab{}.
\newblock \showarticletitle{The probabilistic relevance framework: BM25 and
  beyond}.
\newblock \bibinfo{journal}{\emph{Foundations and Trends{\textregistered} in
  Information Retrieval}} \bibinfo{volume}{3}, \bibinfo{number}{4}
  (\bibinfo{year}{2009}), \bibinfo{pages}{333--389}.
\newblock


\bibitem[Sarvi et~al\mbox{.}(2022)]%
        {sarvi2022understanding}
\bibfield{author}{\bibinfo{person}{Fatemeh Sarvi}, \bibinfo{person}{Maria
  Heuss}, \bibinfo{person}{Mohammad Aliannejadi}, \bibinfo{person}{Sebastian
  Schelter}, {and} \bibinfo{person}{Maarten de Rijke}.}
  \bibinfo{year}{2022}\natexlab{}.
\newblock \showarticletitle{Understanding and Mitigating the Effect of Outliers
  in Fair Ranking}. In \bibinfo{booktitle}{\emph{ACM WSDM}}.
  \bibinfo{pages}{861--869}.
\newblock


\bibitem[Singh and Joachims(2018)]%
        {singh2018fairness}
\bibfield{author}{\bibinfo{person}{Ashudeep Singh} {and}
  \bibinfo{person}{Thorsten Joachims}.} \bibinfo{year}{2018}\natexlab{}.
\newblock \showarticletitle{Fairness of exposure in rankings}. In
  \bibinfo{booktitle}{\emph{ACM SIGIR}}. \bibinfo{pages}{2219--2228}.
\newblock


\bibitem[Singh and Joachims(2019)]%
        {singh2019policy}
\bibfield{author}{\bibinfo{person}{Ashudeep Singh} {and}
  \bibinfo{person}{Thorsten Joachims}.} \bibinfo{year}{2019}\natexlab{}.
\newblock \showarticletitle{Policy learning for fairness in ranking}.
\newblock \bibinfo{journal}{\emph{NeurIPS}}  \bibinfo{volume}{32}
  (\bibinfo{year}{2019}).
\newblock


\bibitem[Singh et~al\mbox{.}(2021)]%
        {singh2021fairness}
\bibfield{author}{\bibinfo{person}{Ashudeep Singh}, \bibinfo{person}{David
  Kempe}, {and} \bibinfo{person}{Thorsten Joachims}.}
  \bibinfo{year}{2021}\natexlab{}.
\newblock \showarticletitle{Fairness in ranking under uncertainty}.
\newblock \bibinfo{journal}{\emph{NeurIPS}}  \bibinfo{volume}{34}
  (\bibinfo{year}{2021}), \bibinfo{pages}{11896--11908}.
\newblock


\bibitem[Sinkhorn and Knopp(1967)]%
        {sinkhorn1967concerning}
\bibfield{author}{\bibinfo{person}{Richard Sinkhorn} {and}
  \bibinfo{person}{Paul Knopp}.} \bibinfo{year}{1967}\natexlab{}.
\newblock \showarticletitle{Concerning nonnegative matrices and doubly
  stochastic matrices}.
\newblock \bibinfo{journal}{\emph{Pacific J. Math.}} \bibinfo{volume}{21},
  \bibinfo{number}{2} (\bibinfo{year}{1967}), \bibinfo{pages}{343--348}.
\newblock


\bibitem[Su et~al\mbox{.}(2022)]%
        {su2022optimizing}
\bibfield{author}{\bibinfo{person}{Yi Su}, \bibinfo{person}{Magd Bayoumi},
  {and} \bibinfo{person}{Thorsten Joachims}.} \bibinfo{year}{2022}\natexlab{}.
\newblock \showarticletitle{Optimizing Rankings for Recommendation in Matching
  Markets}. In \bibinfo{booktitle}{\emph{WWW}}. \bibinfo{pages}{328--338}.
\newblock


\bibitem[Vaswani et~al\mbox{.}(2017)]%
        {vaswani2017attention}
\bibfield{author}{\bibinfo{person}{Ashish Vaswani}, \bibinfo{person}{Noam
  Shazeer}, \bibinfo{person}{Niki Parmar}, \bibinfo{person}{Jakob Uszkoreit},
  \bibinfo{person}{Llion Jones}, \bibinfo{person}{Aidan~N Gomez},
  \bibinfo{person}{{\L}ukasz Kaiser}, {and} \bibinfo{person}{Illia
  Polosukhin}.} \bibinfo{year}{2017}\natexlab{}.
\newblock \showarticletitle{Attention is all you need}.
\newblock \bibinfo{journal}{\emph{NeurIPS}}  \bibinfo{volume}{30}
  (\bibinfo{year}{2017}).
\newblock


\bibitem[Vicol et~al\mbox{.}(2021)]%
        {vicol2021implicit}
\bibfield{author}{\bibinfo{person}{Paul Vicol}, \bibinfo{person}{Jonathan
  Lorraine}, \bibinfo{person}{David Duvenaud}, {and} \bibinfo{person}{Roger
  Grosse}.} \bibinfo{year}{2021}\natexlab{}.
\newblock \showarticletitle{Implicit Regularization in Overparameterized
  Bilevel Optimization}. In \bibinfo{booktitle}{\emph{ICML}}.
\newblock


\bibitem[Wang and Joachims(2021)]%
        {wang2021user}
\bibfield{author}{\bibinfo{person}{Lequn Wang} {and} \bibinfo{person}{Thorsten
  Joachims}.} \bibinfo{year}{2021}\natexlab{}.
\newblock \showarticletitle{User Fairness, Item Fairness, and Diversity for
  Rankings in Two-Sided Markets}. In \bibinfo{booktitle}{\emph{ACM SIGIR}}.
  \bibinfo{pages}{23--41}.
\newblock


\bibitem[Wang et~al\mbox{.}(2018)]%
        {wang2018lambdaloss}
\bibfield{author}{\bibinfo{person}{Xuanhui Wang}, \bibinfo{person}{Cheng Li},
  \bibinfo{person}{Nadav Golbandi}, \bibinfo{person}{Michael Bendersky}, {and}
  \bibinfo{person}{Marc Najork}.} \bibinfo{year}{2018}\natexlab{}.
\newblock \showarticletitle{The lambdaloss framework for ranking metric
  optimization}. In \bibinfo{booktitle}{\emph{ACM CIKM}}.
  \bibinfo{pages}{1313--1322}.
\newblock


\bibitem[Wu et~al\mbox{.}(2022)]%
        {wu2022joint}
\bibfield{author}{\bibinfo{person}{Haolun Wu}, \bibinfo{person}{Bhaskar Mitra},
  \bibinfo{person}{Chen Ma}, \bibinfo{person}{Fernando Diaz}, {and}
  \bibinfo{person}{Xue Liu}.} \bibinfo{year}{2022}\natexlab{}.
\newblock \showarticletitle{Joint Multisided Exposure Fairness for
  Recommendation}.
\newblock \bibinfo{journal}{\emph{ACM SIGIR}} (\bibinfo{year}{2022}),
  \bibinfo{pages}{703--714}.
\newblock


\bibitem[Yadan(2019)]%
        {Yadan2019Hydra}
\bibfield{author}{\bibinfo{person}{Omry Yadan}.}
  \bibinfo{year}{2019}\natexlab{}.
\newblock \bibinfo{title}{Hydra - A framework for elegantly configuring complex
  applications}.
\newblock \bibinfo{howpublished}{Github}.
\newblock
\urldef\tempurl%
\url{https://github.com/facebookresearch/hydra}
\showURL{%
\tempurl}


\bibitem[Yadav et~al\mbox{.}(2021)]%
        {yadav2021policy}
\bibfield{author}{\bibinfo{person}{Himank Yadav}, \bibinfo{person}{Zhengxiao
  Du}, {and} \bibinfo{person}{Thorsten Joachims}.}
  \bibinfo{year}{2021}\natexlab{}.
\newblock \showarticletitle{Policy-gradient training of fair and unbiased
  ranking functions}. In \bibinfo{booktitle}{\emph{ACM SIGIR}}.
  \bibinfo{pages}{1044--1053}.
\newblock


\bibitem[Zehlike et~al\mbox{.}(2017)]%
        {zehlike2017fa}
\bibfield{author}{\bibinfo{person}{Meike Zehlike}, \bibinfo{person}{Francesco
  Bonchi}, \bibinfo{person}{Carlos Castillo}, \bibinfo{person}{Sara Hajian},
  \bibinfo{person}{Mohamed Megahed}, {and} \bibinfo{person}{Ricardo
  Baeza-Yates}.} \bibinfo{year}{2017}\natexlab{}.
\newblock \showarticletitle{Fa* ir: A fair top-k ranking algorithm}. In
  \bibinfo{booktitle}{\emph{ACM CIKM}}. \bibinfo{pages}{1569--1578}.
\newblock


\bibitem[Zehlike and Castillo(2020)]%
        {zehlike2020reducing}
\bibfield{author}{\bibinfo{person}{Meike Zehlike} {and} \bibinfo{person}{Carlos
  Castillo}.} \bibinfo{year}{2020}\natexlab{}.
\newblock \showarticletitle{Reducing disparate exposure in ranking: A learning
  to rank approach}. In \bibinfo{booktitle}{\emph{WWW}}.
  \bibinfo{pages}{2849--2855}.
\newblock


\end{thebibliography}

\end{document}